\documentclass[]{xiaomi}

\usepackage[utf8]{inputenc}
\usepackage[T1]{fontenc}
\usepackage{hyperref}
\usepackage{url}
\usepackage{amsfonts}
\usepackage{nicefrac}
\usepackage{microtype}
\usepackage{xcolor}
\usepackage{overpic}
\usepackage{graphicx}
\usepackage{array}
\usepackage[table]{xcolor}
\usepackage{pifont}
\usepackage{textcomp}
\usepackage{booktabs}
\usepackage{multirow}
\usepackage{subcaption}
\usepackage{amsmath}
\usepackage{amssymb}
\usepackage{ragged2e}
\usepackage{enumitem}
\usepackage{wrapfig}
\usepackage{xspace}

\newcommand{\figref}[1]{Fig.~\ref{#1}}

\newcommand{\mathvec}[1]{\boldsymbol{#1}}
\newcommand{\mathmat}[1]{\mathbf{#1}}

\newcommand{\JWM}{Joint World Model\xspace}

\title{Xiaomi Auto World Model: A Joint World Model Integrating Reconstruction and Generation for Autonomous Driving}

\affiliation{\hyperref[sec:contributors]{Xiaomi Auto World Model Team}}

\abstract{%
This report presents a unified technical system addressing the two core
capabilities of world models for autonomous driving: world representation
and world generation.
For world representation, we propose  \textbf{WorldRec}, a feed-forward reconstruction architecture driven by sparse scene queries. \textbf{WorldRec} initializes structured queries in 3D space, leveraging them to aggregate cross-view, cross-temporal features, thereby naturally enforcing spatial consistency across frames and yielding compact yet high-fidelity 3D Gaussian scene representations.
For world generation, we propose \textbf{WorldGen}, a two-stage training framework of
bidirectional pretraining followed by causal fine-tuning through three progressive
stages (Teacher Forcing, ODE distillation, and DMD), enabling high-quality online
causal video generation in as few as 4 denoising steps.
Building on both modules, we further introduce the \textbf{\JWM}, which deeply
integrates \textbf{WorldRec} and \textbf{WorldGen} to achieve synergistic gains in generation
stability, cross-frame consistency, and visual fidelity, providing a solid
foundation for closed-loop simulation, data synthesis, and end-to-end training
in autonomous driving.%
}

\date{May 2026}
\project{https://JointWM.github.io}

\begin{document}
\thispagestyle{firstheader}
\maketitle

\section{Overview}
  \label{sec:overview}
\begin{figure}[ht]
    \centering
    \includegraphics[width=\linewidth]{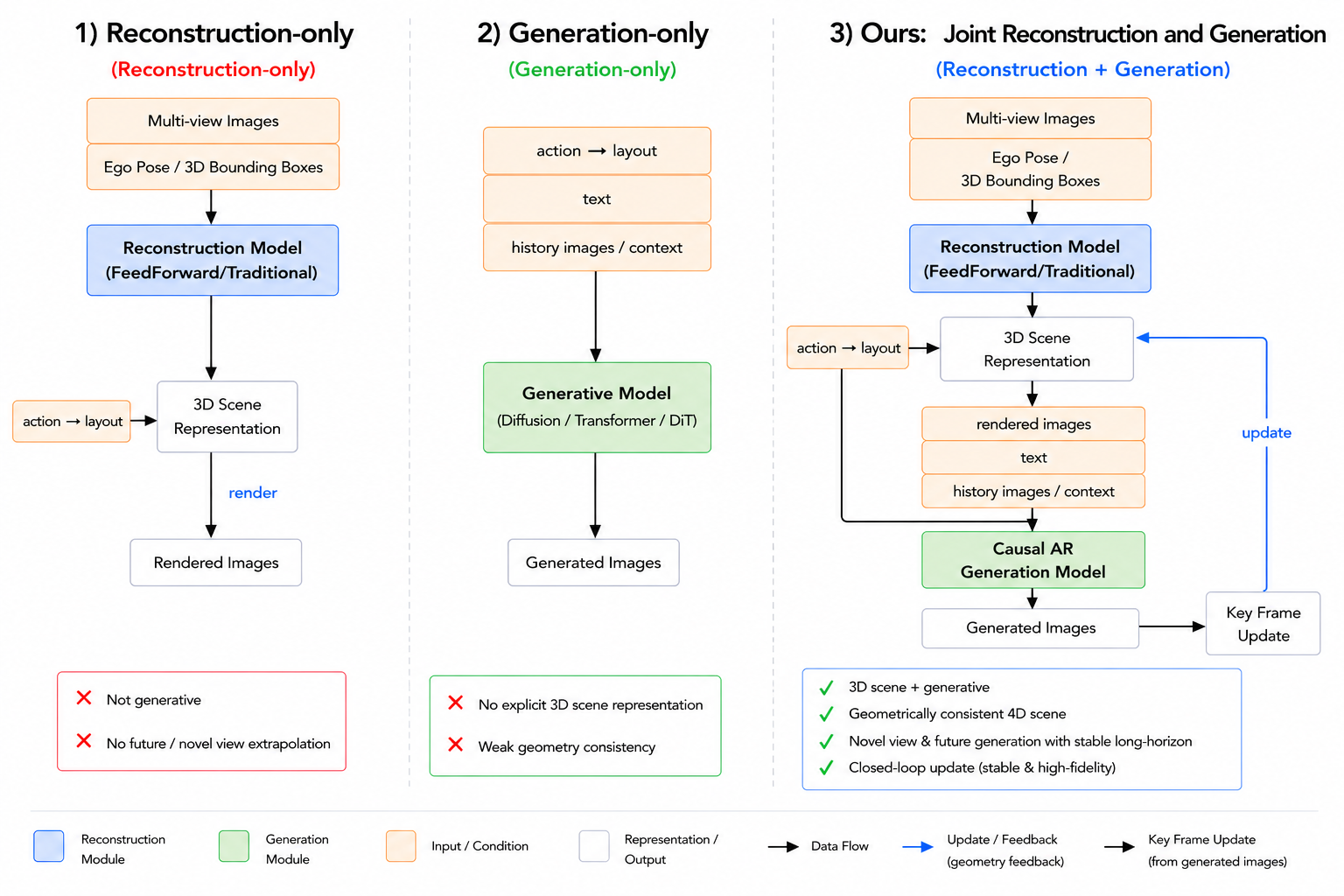}
    \caption{%
     \centering
     Comparison of reconstruction-only, generation-only, and our joint world model.
    }
    \label{fig:teaser}
\end{figure}


World models~\cite{ha2018world} have emerged as a foundational paradigm for autonomous driving, enabling critical capabilities including data synthesis~\cite{dream4drive,genesis} that alleviates the scarcity of long-tail driving scenarios, closed-loop training~\cite{RAD} that facilitates end-to-end policy optimization within differentiable environments, and closed-loop simulation~\cite{yan2024street} that provides high-fidelity virtual evaluation through photorealistic scene rendering. Recent literature has witnessed substantial progress from both the reconstruction perspective
~\cite{yan2024street,chen2024omnire,yang2025storm,chen2025dggt,tan2026ufo} and the generation side~\cite{gaia1,drivedreamer,magicdrive,gaia2,genesis}, 
converging toward a hybrid reconstruction-generation paradigm. This paradigm typically proceeds in two stages: first constructing a 3D scene representation~\cite{kerbl3Dgaussians} from multi-view observations, then leveraging the resulting geometric prior to condition video generation models~\cite{ho2020denoising,peebles2023scalable}. While such an approach has demonstrated promising properties---notably spatiotemporal consistency and precise viewpoint controllability---current instantiations remain limited in scalability, real-time applicability, and the depth of integration between the two stages.

We identify three principal bottlenecks. \textit{First}, on the representation side, conventional per-scene optimization of 3D Gaussian primitives entails multi-hour training per sequence and exhibits poor generalization to novel viewpoints. Although recent feed-forward methods~\cite{yang2025storm,chen2025dggt,tan2026ufo} mitigate the optimization cost, their reliance on Dense Prediction Transformer (DPT) heads for pixel-aligned Gaussian prediction introduces systematic deficiencies: independently predicted per-frame Gaussians inevitably produce ghosting artifacts and layered surface duplications upon concatenation in 3D space, while the primitive count scales to hundreds of millions per clip, imposing prohibitive rendering overhead. \textit{Second}, on the generation side, causal diffusion models trained \textit{ab initio} lack the expressive scene priors necessary for high-fidelity synthesis; the requirement of hundreds of denoising steps at inference precludes real-time deployment; and the well-documented exposure bias~\cite{ranzato2016sequence} inherent to autoregressive generation induces progressive content drift over extended horizons. \textit{Third}, the coupling between reconstruction and generation in existing systems remains superficial---most approaches treat the two as disjoint modules, rendering it difficult to reconcile the geometric fidelity demanded by precise scene representation with the distributional diversity essential to generative prediction. Moreover, recent hybrid efforts such as NeoVerse~\cite{neoverse}, while demonstrating compelling results for general 4D scene understanding from monocular videos, lack the domain-specific architectural considerations required by autonomous driving, including multi-camera geometric consistency, ego-motion conditioning, and structured layout control.

This report presents a unified technical framework that systematically addresses the aforementioned challenges through three tightly integrated components. \textbf{WorldRec} formulates scene reconstruction as a sparse-query aggregation problem: a compact set of learnable 3D queries are initialized in world space and progressively enriched via cross-view, cross-temporal feature fusion with visibility-aware weighting, yielding geometrically consistent Gaussian scene representations that reduce reconstruction time from hours to approximately 10 seconds per clip while eliminating the ghosting and redundancy artifacts of prior feed-forward approaches. \textbf{WorldGen} introduces a two-stage training curriculum built upon a Diffusion Transformer (DiT) backbone conditioned on heterogeneous signals including ego trajectory, camera parameters, layout maps, and free-form text. Bidirectional pretraining with unrestricted temporal attention first acquires rich spatiotemporal scene priors; the model is subsequently converted to an autoregressive generator through progressive causal fine-tuning comprising Teacher Forcing for causal adaptation, ODE distillation for step reduction ($50 \to 4$ steps, yielding ${\sim}12\times$ inference acceleration), and Distribution Matching Distillation (DMD) for mitigating exposure bias---collectively enabling stable generation of up to one-minute driving videos at 0.19\,s/frame. The \textbf{\JWM} achieves deep integration of both modules through complementary architectural extensions: an incremental scene fusion mechanism allows WorldRec to maintain and progressively expand a globally consistent 4D Gaussian representation as new observations arrive, while WorldGen is augmented with an ego-projected rendered-prior conditioning pathway that provides a coarse geometric scaffold for generation in unobserved regions while preserving photometric consistency where reconstruction coverage exists. Within this tightly coupled framework, the deterministic geometric constraints from WorldRec suppress error accumulation and content drift during long-horizon autoregressive generation, while the generative capacity of WorldGen compensates for spatially and temporally unreconstructed regions, yielding synergistic improvements along three axes: temporal stability, cross-view consistency, and visual fidelity.

As illustrated in Figure~\ref{fig:teaser}, reconstruction-only methods achieve geometrically precise scene recovery yet remain fundamentally confined to observed data, lacking the capacity to synthesize future or unseen viewpoints. Conversely, generation-only approaches model scene dynamics and afford controllable prediction, but the absence of an explicit 3D representation leads to weak geometric consistency and accumulated drift over long horizons. Our joint formulation unifies the complementary strengths of both paradigms: a compact 3D scene state serves as a structured geometric anchor for the generative process, while the generation model extends predictive capability beyond the observation boundary. The resulting closed-loop architecture establishes a principled foundation for closed-loop simulation, synthetic data generation, and end-to-end training in autonomous driving.

\begin{table}[t]
\centering
\captionsetup{justification=centering}
\caption{Comparison of different world modeling paradigms.}
\small
\begin{tabular}{lccccc}
\toprule
 & Recon-only & Gen-only & Neoverse & AlpaDreams & \textbf{Ours} \\
\midrule
Explicit 3D Scene & \ding{51} & \ding{55} & \ding{51} & \ding{55} & \ding{51} \\

Generative Capability & \ding{55} & \ding{51} & \ding{51} & \ding{51} & \ding{51} \\

Novel View Synthesis & \ding{51} & \ding{51} & \ding{51} & \ding{51} & \ding{51} \\

Future Prediction & \ding{55} & \ding{51} & \ding{51} & \ding{51} & \ding{51} \\

Geometry Consistency & Strong & Weak & Medium & Weak & Strong \\

Long-horizon Stability & Static & Drift & Medium & Medium & Stable \\

\bottomrule
\end{tabular}
\end{table}

\section{Related Work}
\label{sec:related}

\subsection{3D Scene Reconstruction.}
3D scene reconstruction aims to recover accurate geometric structures and achieve photorealistic appearance rendering from multi-view observations. Early approaches relied on explicit representations such as point clouds and meshes, which struggle to capture fine surface details and view-dependent appearance. 3D Gaussian Splatting (3DGS) \cite{kerbl3Dgaussians} has since emerged as the dominant paradigm, representing scenes as collections of explicit 3D Gaussian primitives rendered via efficient differentiable rasterization, offering a compelling combination of geometric expressiveness, rendering speed, and visual fidelity.

Early work adapts 3DGS to driving scenes through iterative per-scene fitting. StreetGaussians \cite{yan2024street} decomposes the scene into a static background and dynamic foreground vehicles, modeling each dynamic actor as a rigid group of Gaussian primitives with optimizable poses and 4D spherical harmonics to capture appearance variation across frames. DrivingGaussian \cite{zhou2024drivinggaussian} extends this with composite dynamic Gaussian graphs for multi-object modeling. To address non-rigid actors overlooked by prior work, OmniRe \cite{chen2024omnire} builds dynamic neural scene graphs on 3DGS and constructs multiple canonical-space Gaussian representations covering vehicles, pedestrians, and cyclists, including skinned LBS models for deformable bodies, enabling holistic reconstruction of all dynamic actors at ~60Hz simulation throughput. S³Gaussian \cite{huang2024s3gaussian} further explores self-supervised static-dynamic decomposition without requiring costly 3D bounding box annotations. Uni-Gaussians \cite{yuan2025unigaussians} unifies camera and LiDAR rendering within a single 3DGS framework, enabling joint optimization across heterogeneous sensor modalities. ExtraGS \cite{tan2025extrags} introduces Difix3D \cite{wu2025difix3d+} as a novel-view data augmentation strategy, substantially improving rendering quality across lane changes and large viewpoint shifts. 3DGUT \cite{wu2024dgut} and ParkGaussian \cite{wei2025parkgaussian} address the specific challenge of fisheye camera reconstruction in autonomous driving, extending 3DGS to handle severe radial distortion inherent to wide field-of-view optics. Despite achieving high reconstruction quality, these per-scene optimization methods suffer from prohibitive training costs that require tens of minutes to hours per new sequence, and struggle to generalize to novel viewpoints, fundamentally limiting their scalability.

To overcome these limitations, feedforward models learn generalizable priors from large-scale data and infer Gaussian representations in a single forward pass. PixelSplat \cite{charatan2024pixelsplat} and MVSplat \cite{chen2024mvsplat} pioneer this paradigm with epipolar and cost-volume based depth estimation, but struggle with the low camera overlap and unbounded dynamics of driving environments. STORM \cite{yang2025storm} is the first feedforward 3DGS framework targeting large-scale outdoor dynamic scenes, predicting per-frame Gaussians and scene flows from sparse context frames via learnable motion tokens, recovering scene dynamics without explicit motion supervision. UFO \cite{tan2026ufo} builds upon STORM by incorporating a temporal fusion module that aggregates multi-frame context representations, further improving reconstruction consistency over longer sequences. DGGT \cite{chen2025dggt} further removes the dependency on known camera calibration, reformulating pose as a model output for pose-free dynamic reconstruction from unposed images, while introducing a lifespan head to modulate Gaussian visibility for temporal consistency across long sequences.

\subsection{Video Generation for Autonomous Driving}

Recent advances in generative world models have demonstrated the feasibility of synthesizing realistic driving videos from large-scale real-world data.
GAIA-1~\cite{gaia1} pioneered this direction by combining a world-model backbone with a video diffusion decoder.
Subsequent works have improved controllability and scalability along multiple axes.
DriveDreamer~\cite{drivedreamer} introduces structured layout conditioning,
while Drive-WM~\cite{drivewm} incorporates multi-view forecasting for planning-aware generation.
MagicDrive~\cite{magicdrive} and its extension MagicDrive-V2~\cite{magicdriveV2} enable controllable 3D geometry and long-form high-resolution synthesis.
Panacea~\cite{panacea} focuses on panoramic multi-view consistency,
and GAIA-2~\cite{gaia2} unifies fine-grained conditioning and cross-camera consistency within a latent diffusion framework.

Recent works move beyond RGB video synthesis to explore richer world representations and scalable training paradigms.
WorldSplat~\cite{worldsplat} adopts feed-forward 3D Gaussian representations for multi-view generation,
while OccWorld~\cite{occworld} models dynamic scenes in 3D occupancy space.
GenAD~\cite{genad} and Cosmos-Drive-Dreams~\cite{cosmosdrive} leverage large-scale driving corpora to pre-train generative models,
and Dream4Drive~\cite{dream4drive} demonstrates the effectiveness of synthetic data f
or closed-loop training.

Despite this progress, three key challenges remain.
First, training causal diffusion models from scratch often lacks strong generative priors, limiting sample quality.
Second, the reliance on hundreds of denoising steps prevents real-time deployment.
Third, exposure bias in autoregressive generation leads to long-horizon drift and hallucination.
To address these issues, we propose \textbf{WorldGen}, a two-stage framework that combines bidirectional pre-training with progressive causal fine-tuning
(Teacher Forcing $\rightarrow$ ODE distillation $\rightarrow$ DMD),
enabling stable and high-quality online video generation with as few as four denoising steps.

\subsection{Joint Reconstruction and Generation}

Reconstruction and generation are two complementary pillars of world models, yet most existing systems treat them as separate modules. Reconstruction-only methods, such as per-scene 3DGS optimization~\cite{yan2024street,chen2024omnire}, excel at geometrically precise scene rendering but lack the ability to synthesize unseen regions. Conversely, generation-only methods like GAIA-2~\cite{gaia2} can imagine beyond observed data but often suffer from geometric drift and cross-frame inconsistency.

Recent efforts have begun to bridge this gap. WorldSplat~\cite{worldsplat} generates multi-view scenes with feed-forward 3D Gaussian representations, while GAIA-2~\cite{gaia2} unifies reconstruction and generation within a single latent diffusion framework. More recently, NeoVerse~\cite{neoverse} proposes a reconstruction-generation hybrid architecture trained on one million in-the-wild monocular videos, achieving impressive generalization across diverse domains. However, existing approaches including NeoVerse primarily focus on static scene reconstruction or unbounded trajectory generation, lacking dedicated designs for autonomous driving's unique requirements: multi-camera rigs with fixed extrinsic calibration, ego-motion awareness, and the need for long-horizon causal prediction under structured layout conditions.

To address these domain-specific challenges, we propose a \textbf{Joint World Model} that deeply integrates WorldRec and WorldGen. Within this framework, the deterministic geometric constraints from WorldRec suppress generative drift in long-horizon autoregressive generation, while the rich imagination of WorldGen compensates for unreconstructed regions. Unlike NeoVerse which targets general 4D scene understanding from arbitrary monocular videos, our system is specifically architected for autonomous driving, leveraging multi-view temporal consistency, layout conditioning, and efficient 4-step causal generation to achieve synergistic gains in stability, consistency, and fidelity for closed-loop simulation and end-to-end training.
\section{Method}
\label{sec:methods}

\begin{figure}[h]
    \centering
    \includegraphics[width=\linewidth]{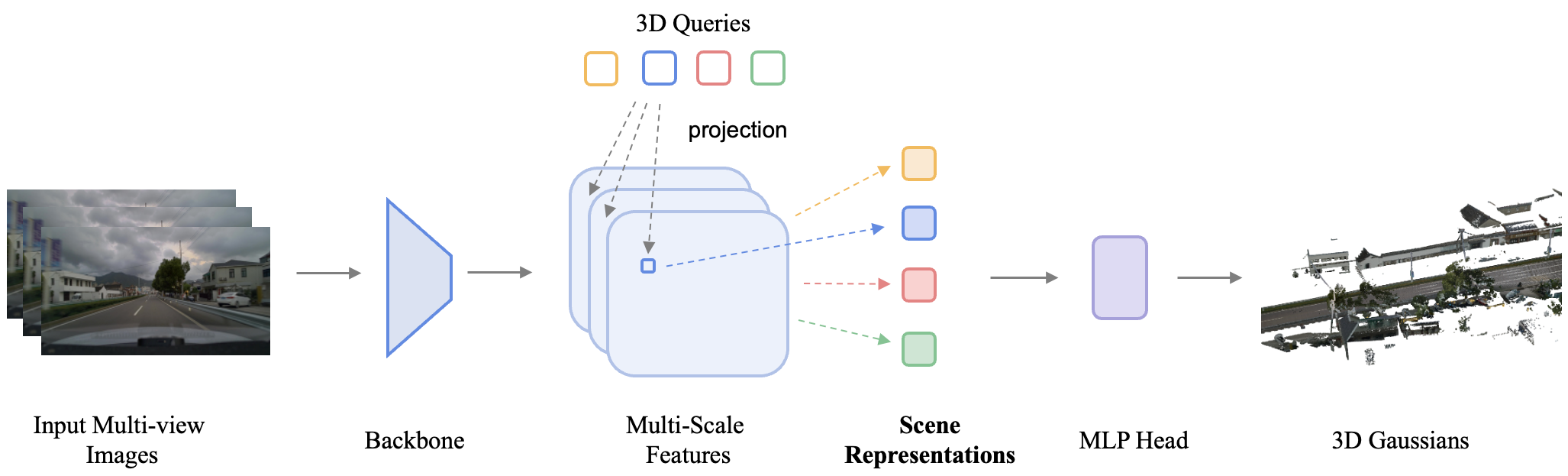}
    \caption{%
     \centering
     WorldRec Network Architecture
    }
    \label{fig:worldrec}
\end{figure}

\subsection{World Representation}
\label{sec:worldrec}

World representation is a cornerstone of world modeling, among which 3D Gaussian Splatting (3DGS) \cite{kerbl3Dgaussians} has become a prevailing choice owing to its explicit geometry and real-time rendering capability. Traditional 3DGS reconstruction, however, depends on per-scene optimization that is time-consuming and difficult to scale. To overcome this limitation, feedforward 3DGS has rapidly emerged as the mainstream paradigm. Nevertheless, most existing feedforward approaches \cite{yang2025storm,chen2025dggt,tan2026ufo} rely on a Dense Prediction Transformer (DPT) head to predict pixel-aligned 3D Gaussians, a design that suffers from several fundamental limitations. Since each input image independently produces its own set of Gaussians, naive concatenation of per-frame predictions in 3D space inevitably results in ghosting artifacts and layered duplications along object surfaces. Moreover, each frame generates hundreds of thousands of Gaussians, accumulating to potentially hundreds of millions across a single clip, imposing severe redundancy and rendering overhead that scales poorly with sequence length.
Motivated by these observations, we propose to represent the scene using a compact set of sparse tokens rather than dense, pixel-aligned primitives. Each token aggregates features from multi-view, multi-temporal image observations through feature fusion, enabling the model to learn holistic, view-consistent scene representations. By grounding each Gaussian in information observed across multiple viewpoints and timestamps, our formulation explicitly enforces multi-view consistency, substantially mitigating the ghosting and layering artifacts inherent to per-frame pixel-aligned prediction paradigms.

\subsubsection{Network Architecture}
As shown in Figure~\ref{fig:worldrec}, given a multi-camera, multi-temporal image
sequence $\{\mathmat{I}^t_c\}$ from a driving scene (where $c$ is the camera
index and $t$ is the time index), \textbf{WorldRec} builds a sparse Gaussian scene
representation through the following pipeline.

\paragraph{Multi-scale feature extraction.}
All input images are processed by a shared-weight visual backbone to extract
multi-scale feature maps $\{\mathmat{F}^t_{c,l}\}$, where $l$ denotes the
feature scale.
The multi-scale design enables the network to jointly perceive fine-grained
texture details and large-scale semantic structures, providing a rich feature
basis for subsequent cross-view aggregation.

\paragraph{3D query initialization and projection sampling.}
We initialize $N$ sparse 3D queries in world space, each representing a
reference coordinate $\mathvec{p} = [X, Y, Z]^\top$ that encodes a query for
Gaussian primitive attributes at that spatial location.
For each query, the reference point is projected onto the feature map of the
$c$-th view at scale $l$ using calibrated intrinsics and extrinsics, yielding
the 2D sampling coordinate:
\begin{equation}
  \mathvec{u}^{c,l} = \pi_c(\mathvec{p}).
  \label{eq:projection}
\end{equation}
Local image features are then extracted via bilinear interpolation:
\begin{equation}
  \mathvec{f}^{c,l} =
    \operatorname{BilinearInterp}\!\left(
      \mathmat{F}^t_{c,l},\;\mathvec{u}^{c,l}
    \right).
  \label{eq:bilinear}
\end{equation}
Each 3D query thus corresponds to a set of sampled features
$\{\mathvec{f}^{c,l}\}$ across all visible views and feature scales.

\paragraph{Cross-view cross-temporal feature aggregation.}
Since the same spatial location may be observed simultaneously by multiple
cameras and is subject to motion and occlusion changes across frames, effective
fusion across views and time steps is critical.
We employ a visibility-aware weighted aggregation module to fuse the sampled
multi-view features $\{\mathbf{f}^{c,l}\}$. Concretely, a lightweight network
predicts a normalized weight $w^{c,l}$ for each sampled feature based on its
visibility and reliability, and the final scene token for query $i$ is obtained
via weighted average:
\begin{equation}
  \mathbf{q}_i = \sum_{c,l} w^{c,l}\, \mathbf{f}^{c,l}_i
  \label{eq:aggregation}
\end{equation}
The fusion weights are determined by the estimated visibility and feature
quality of each view, enabling the model to adaptively emphasize informative,
less-occluded observations and suppress noisy features caused by occlusion or
surface reflections. In the temporal dimension, features from adjacent frames
are aligned before aggregation, further enhancing consistency in dynamic scenes.

\paragraph{Gaussian attribute decoding.}
A lightweight MLP head maps each aggregated scene token $\mathvec{q}_i$ to the
full attribute set of the corresponding 3D Gaussian primitive---spatial position
offset $\Delta\mathvec{p}$, color $\mathvec{c}$ (RGB), opacity $\alpha$,
covariance scale $\mathvec{s}$, and rotation quaternion $\mathvec{r}$:
\begin{equation}
  \left(
    \Delta\mathvec{p}_i,\;\mathvec{c}_i,\;\alpha_i,\;\mathvec{s}_i,\;\mathvec{r}_i
  \right) = \operatorname{MLP}(\mathvec{q}_i).
  \label{eq:mlp}
\end{equation}
The final Gaussian center is given by
$\hat{\mathvec{p}}_i = \mathvec{p}_i + \Delta\mathvec{p}_i$, allowing the model
to retain the prior spatial layout while enabling fine-grained localization.

\paragraph{Rendering and supervision.}
Based on all 3D Gaussian primitives, differentiable Gaussian
rasterization~\cite{kerbl3Dgaussians} renders arbitrary target views, producing
rendered images $\hat{\mathmat{I}}_c$.
During training, rendered results and corresponding ground-truth images
$\mathmat{I}_c$ jointly form the supervision signal; the loss function combines
pixel-level reconstruction accuracy and perceptual quality:
\begin{equation}
  \mathcal{L} = \mathcal{L}_{\text{pixel}}
               + \lambda\,\mathcal{L}_{\text{perceptual}}.
  \label{eq:rec-loss}
\end{equation}
Applying rendering supervision across all camera views simultaneously explicitly
guides the model to learn cross-view-consistent scene geometry and appearance,
reinforcing the spatial coherence of the sparse representation.

\subsection{World Generation}
\label{sec:worldgen}


We adopt the Diffusion Transformer (DiT)~\cite{peebles2023scalable} as the
backbone of our world generation module.
DiT partitions video frame sequences into patch-level tokens and models their
joint spatiotemporal dynamics through Transformer self-attention.
Compared with U-Net-based diffusion architectures, DiT offers superior
long-sequence modeling capacity, more flexible multi-modal conditioning, and
favorable scaling behavior, making it the prevailing paradigm in high-fidelity
video generation and a common choice in recent driving scene
generators~\cite{genesis,drivelaw,dream4drive}.

\begin{figure}[t]
    \centering
    \includegraphics[width=\linewidth]{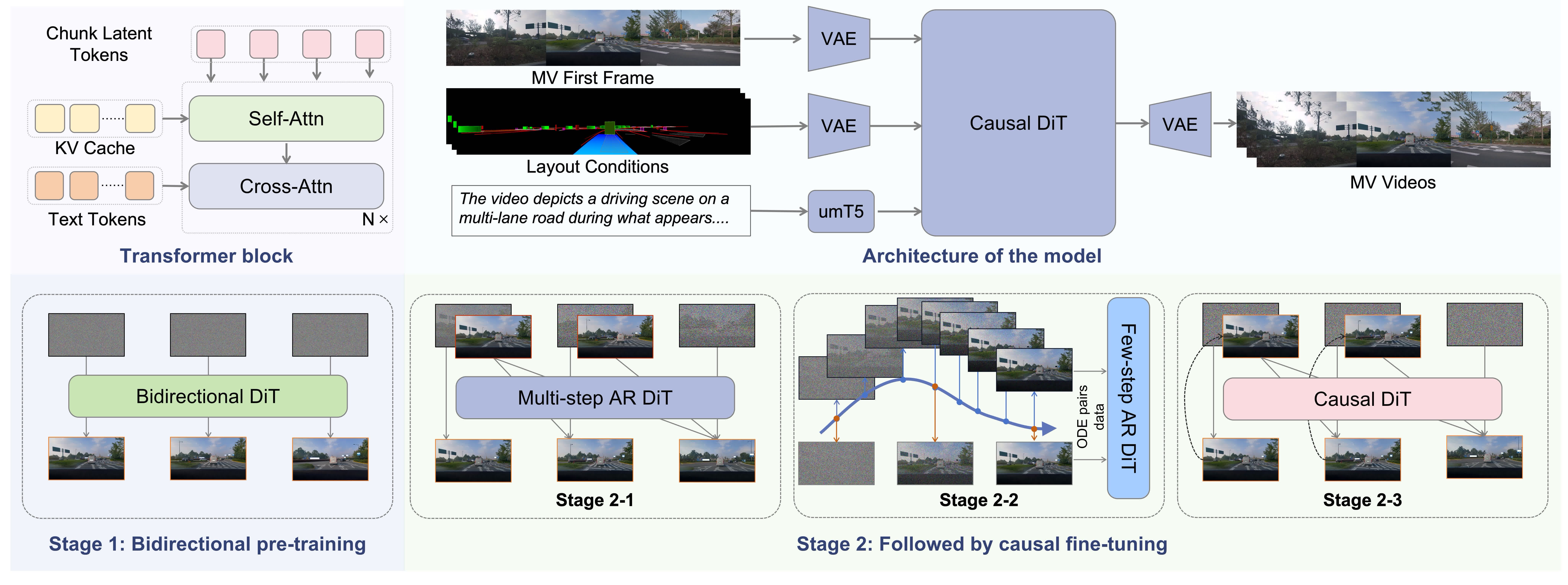}
    \caption{%
      \textbf{WorldGen architecture and two-stage training framework:}
      Top left: Transformer blocks of the causal DiT. Top right: multi-view frames, layout conditions, and text prompts are encoded into a shared latent space using modality-specific encoders (VAE for vision, umT5 for language). The fused representations are iteratively denoised by a causal DiT to generate multi-view video outputs. Bottom: bidirectional pre-training followed by causal fine-tuning.
    }
    \label{fig:worldgen-stages}
\end{figure}

As illustrated in Figure~\ref{fig:worldgen-stages}, the proposed architecture
integrates heterogeneous conditioning signals to ensure fine-grained
controllability over the generation process.
Concretely, the \textbf{multi-view first frame} and \textbf{layout condition}
are projected into clean and conditional latent representations, respectively,
via a Variational Autoencoder (VAE), while free-form textual descriptions are
embedded using a pre-trained multilingual language model (umT5).
These multi-modal features are concatenated along the token dimension and
injected into the \textbf{Causal DiT}, which performs iterative latent-space
denoising to synthesize temporally coherent, high-fidelity
\textbf{multi-view videos}.
Our training strategy, termed \textbf{WorldGen}, follows a
\emph{bidirectional pre-training $\to$ causal fine-tuning} curriculum
(Figure~\ref{fig:worldgen-stages}):

\begin{itemize}[leftmargin=*, itemsep=0.3em]
  \item \textbf{Stage~1 --- Bidirectional pre-training.}
        We first train a DiT with full bidirectional temporal attention under
        the standard denoising diffusion objective.
        Unrestricted access to the complete temporal context allows the model
        to capture the global spatiotemporal distribution of driving scenes,
        thereby establishing a strong generative prior.

  \item \textbf{Stage~2 --- Causal fine-tuning.}
        We subsequently impose a causal attention mask to convert the model into
        an autoregressive generator suitable for streaming inference.
        The causal model is then progressively refined through three
        stages---Teacher Forcing, ODE distillation, and
        Distribution Matching Distillation (DMD)---to achieve high-quality,
        low-latency online generation.
\end{itemize}

The rationale underlying this two-stage design is twofold.
First, bidirectional models---having access to the full temporal
context---learn data distributions more sample-efficiently and develop richer
scene priors than their causal counterparts trained from scratch.
Second, causal models are inherently suited to streaming generation and
long-horizon temporal extrapolation, properties that are indispensable for
real-time simulation.
By initializing the causal model from a converged bidirectional checkpoint, we
circumvent the well-documented optimization difficulties of training causal
diffusion models \emph{ab initio}, while fully inheriting the expressive scene
priors accumulated during pre-training.

\subsubsection{Bidirectional Model Training}
\label{sec:bi-training}

The first stage aims to establish a base model with strong scene generation
capability.
The model employs full bidirectional temporal attention, in which every token
attends freely to all tokens across the entire temporal extent without any
causal masking.
This unrestricted attention pattern enables the model to exploit both past and
future context simultaneously, leading to more sample-efficient learning of
the joint spatiotemporal distribution of driving scenes.

The model is conditioned on ego trajectory, camera intrinsics, and camera
extrinsics, which are encoded and injected into each DiT layer as control
signals.
Instead of diffusion-based denoising, we adopt a rectified flow formulation
to model the continuous transformation from noise to data.
Specifically, we construct a linear interpolation between a noise sample 
$z \sim \mathcal{N}(0, I)$ and a data sample $x_0$:
\begin{equation}
  x_t = (1 - t)\,z + t\,x_0,
  \quad t \sim \mathcal{U}(0,1),
  \label{eq:rf-interp}
\end{equation}

and train a velocity field $v_\theta$ to match the ground-truth flow:
\begin{equation}
  v^*(x_t, t) = \frac{d x_t}{dt} = x_0 - z.
\end{equation}

The model is optimized via the flow matching objective:
\begin{equation}
  \mathcal{L}_{\text{rf}} =
    \mathbb{E}_{t,\,x_0,\,z}
    \left[
      \left\| v_\theta(x_t,\,t,\,c) - (x_0 - z) \right\|_2^2
    \right],
  \label{eq:rf-loss}
\end{equation}
where $c$ denotes the aggregated conditioning signals.
At inference time, generation is performed by integrating the learned velocity
field from noise to data along the rectified flow trajectory.

\subsubsection{Causal Model Training}
\label{sec:causal-training}

The bidirectional model requires processing all frames simultaneously at
inference and therefore cannot support frame-by-frame autoregressive generation,
failing to meet the demands of online streaming inference and long-horizon
extrapolation in closed-loop simulation.
We migrate the bidirectional model to a causal generator through three
progressive training stages~\cite{zhu2026causal}, each targeting a distinct challenge.

\paragraph{Teacher Forcing.}
Teacher Forcing replaces the bidirectional model's full temporal attention with
a causal attention mask, enforcing that each frame can only attend to current noisy frame
and past clean frames:
\begin{equation}
  M_{ij} = \begin{cases}
    0        & \text{if } j \leq i, \\
    -\infty  & \text{if } j > i,
  \end{cases}
  \label{eq:causal-mask}
\end{equation}
where $i, j$ are the temporal indices of the query and key frames.
Model parameters are warm-started from the pretrained bidirectional model.
During training, \textbf{ground-truth frames} serve as historical context,
and the model learns to predict the current frame conditioned on the true past:
\begin{equation}
  \mathcal{L}_{\text{TF}} =
    \mathbb{E}_{t,\,\epsilon}
    \!\left[\left\|\epsilon_\theta\!\left(
      x_t^{(i)},\;t,\;c,\;x_{\text{GT}}^{(<i)}
    \right) - \epsilon\right\|_2^2\right],
  \label{eq:tf-loss}
\end{equation}
where $x_{\text{GT}}^{(<i)}$ denotes the ground-truth context before frame $i$.
This strategy is training-stable and converges rapidly because the context is
always drawn from clean ground truth, avoiding error propagation.
However, Teacher Forcing introduces the classical \textbf{exposure bias}
problem~\cite{ranzato2016sequence}: at inference time the model must condition
on its own previously generated frames rather than ground truth, and this
train--inference distribution mismatch accumulates with autoregressive steps,
causing quality degradation and content drift in long-horizon generation.

\paragraph{ODE Distillation.}
After Teacher Forcing, the model still requires approximately 50 denoising steps
at inference, incurring high computational cost that is incompatible with
real-time simulation.
ODE distillation leverages the trajectory consistency of deterministic ODE
solvers~\cite{song2020denoising} to train a student model to match 50-step
sampling quality with only \textbf{4 steps}, improving inference efficiency
by $\sim$12$\times$.
The probability-flow ODE corresponding to the diffusion model is:
\begin{equation}
  \frac{dx_t}{dt} = f_\theta(x_t,\,t,\,c).
  \label{eq:pf-ode}
\end{equation}
The teacher model obtains high-quality samples via $N\!=\!50$ steps:
$\hat{x}_0^{\text{teacher}} = \operatorname{ODESolve}(x_T, f_\theta, N\!=\!50)$.
The distillation objective trains the student $f_\phi$ with $K\!=\!4$ steps:
\begin{equation}
  \mathcal{L}_{\text{ODE}} =
    \mathbb{E}_{x_T}
    \!\left[\left\|f_\phi(x_T,\;K\!=\!4)
    - \operatorname{sg}\!\left[\hat{x}_0^{\text{teacher}}\right]
    \right\|_2^2\right],
  \label{eq:ode-loss}
\end{equation}
where $\operatorname{sg}[\cdot]$ denotes the stop-gradient operation.
The deterministic, unique nature of probability-flow ODE trajectories provides
stable and consistent supervision signals, making them more suitable for
distillation than stochastic SDE-based sampling.
This efficiency gain is critical for real-world deployment, enabling the causal
generation model to approach real-time frame rates.

\paragraph{DMD.}
DMD (Distribution Matching Distillation)~\cite{yin2024improved} directly
addresses the exposure bias introduced by Teacher Forcing.
In the DMD stage, historical context frames are replaced by the model's
\textbf{own generated outputs} during training, exposing the model to its own
generation distribution and substantially closing the train--inference
distribution gap.
Specifically, historical context is first generated by the model itself:
\begin{equation}
  \hat{x}^{(<i)} = G_\phi(\epsilon,\,c),
  \quad \epsilon \sim \mathcal{N}(0,I).
  \label{eq:self-gen}
\end{equation}
The DMD training objective combines a denoising regression loss and a
distribution matching loss:
\begin{equation}
  \mathcal{L}_{\text{DMD}}
  = \underbrace{
      \mathbb{E}_{t,\,\epsilon}\!\left[\left\|\epsilon_\phi\!\left(
        x_t^{(i)},\;t,\;c,\;\hat{x}^{(<i)}
      \right) - \epsilon\right\|_2^2\right]
    }_{\mathcal{L}_{\text{reg}}:\;\text{denoising regression}}
  + \lambda\;\underbrace{
      D_{\mathrm{KL}}\!\left(p_\phi\;\|\;p_{\text{data}}\right)
    }_{\mathcal{L}_{\text{dist}}:\;\text{distribution matching}},
  \label{eq:dmd-loss}
\end{equation}
where $\hat{x}^{(<i)}$ is the model's own generated preceding frames (replacing
ground truth) and $\lambda$ is a balancing coefficient.
Compared to sample-level regression losses, distribution matching more
effectively captures the global characteristics of the generation distribution,
avoiding the blurring associated with per-frame alignment.
After DMD training, the causal model achieves significantly improved temporal
stability and content consistency in long-horizon autoregressive generation,
effectively suppressing error accumulation and content drift.
Combined with 4-step efficient sampling from ODE distillation, this achieves a
favorable balance among generation quality, temporal stability, and inference
speed.

\subsection{Joint World Model}
\label{sec:joint}

\begin{figure}[htbp]
    \centering
    \includegraphics[width=\linewidth]{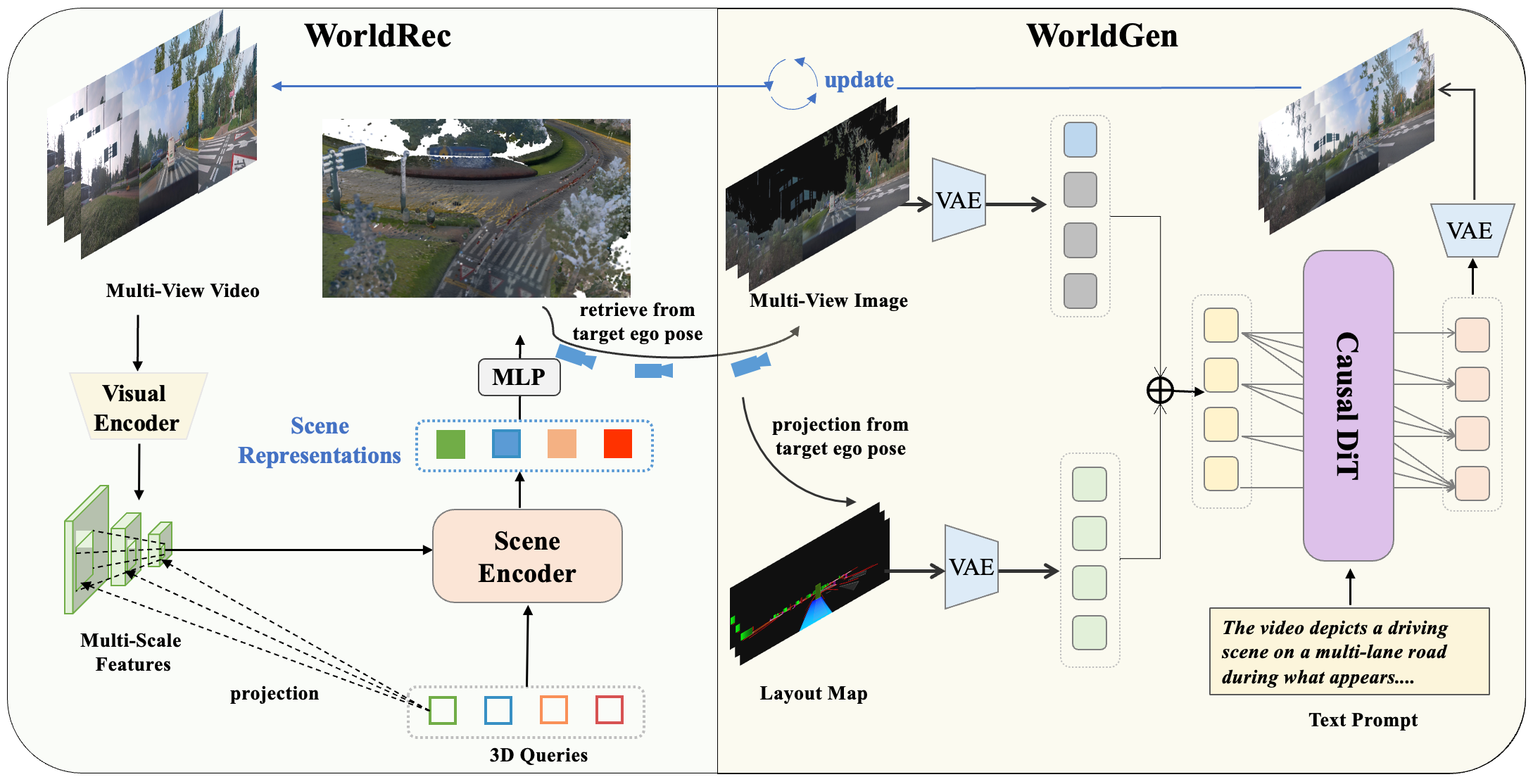}
    \caption{%
    \centering
      \JWM Architecture
    }
    \label{fig:joint-arch}
\end{figure}

High-quality world modeling requires two complementary capabilities: precise
understanding and compact representation of observed scenes, and generative
prediction of unobserved states.
Neither capability alone is sufficient---reconstruction-only methods lack the
imagination to synthesize unseen regions, while generation-only methods cannot
guarantee consistency with known scene content.
We therefore propose the \textbf{\JWM}, which deeply integrates \textbf{WorldRec} and
\textbf{WorldGen} to achieve synergistic gains across both dimensions.

\paragraph{Contribution of world representation.}
The sparse scene tokens produced by \textbf{WorldRec} capture scene geometry,
appearance texture, and temporal dynamics with minimal redundancy, forming a
compact yet information-rich scene prior.
This structured 4D representation provides \textbf{WorldGen} with reliable spatial
anchors, effectively preventing geometric drift and cross-frame inconsistency
during generation.

\paragraph{Contribution of world generation.}
The causal generation model, conditioned on the sparse scene representation,
performs reasonable extrapolation and inpainting of viewpoints, occluded
regions, and future time steps beyond the observation boundary, endowing the
world model with genuine generative imagination.

\begin{figure}[t]
    \centering
    \begin{subfigure}[b]{0.48\linewidth}
        \centering
        \includegraphics[width=\linewidth]{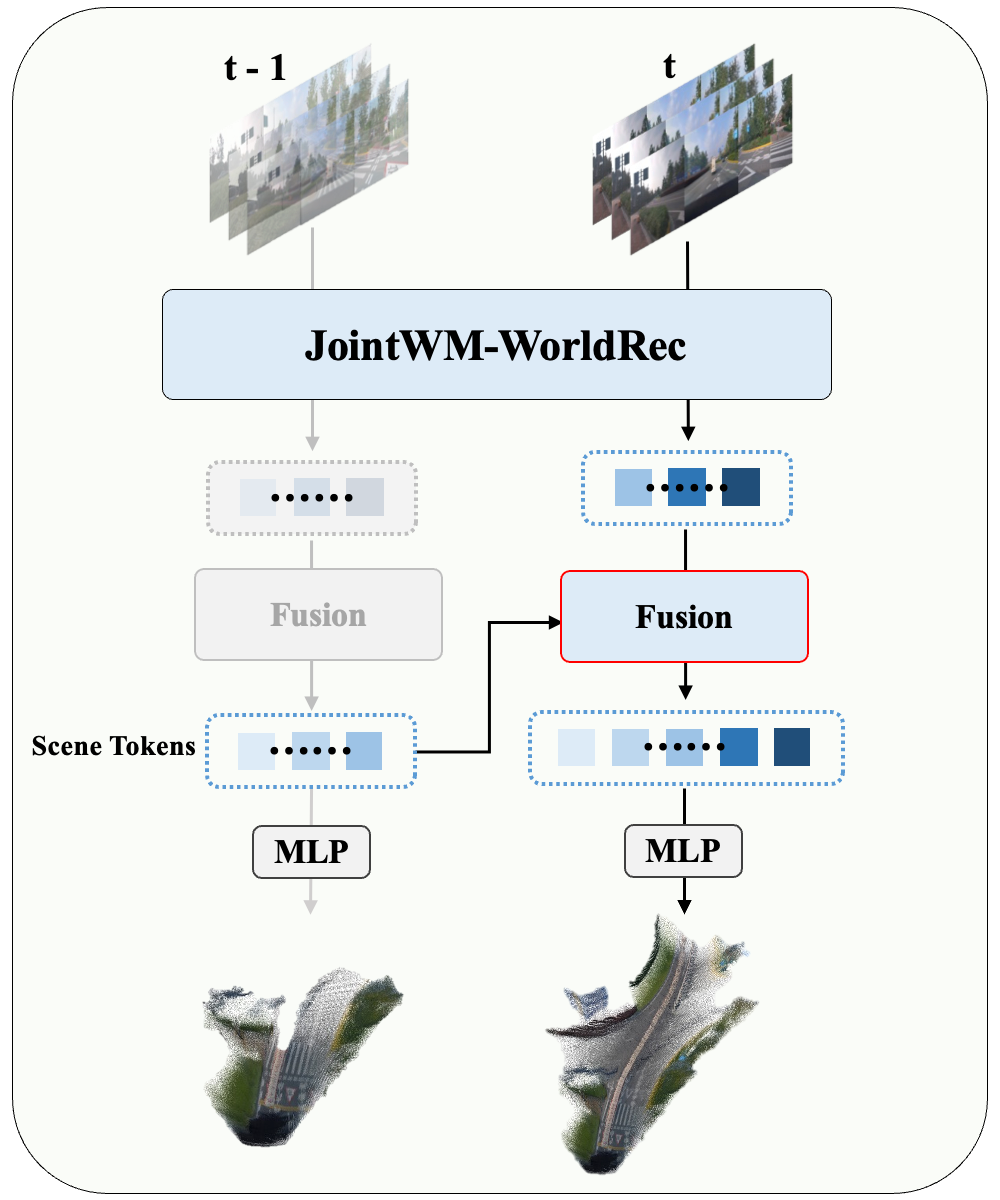}
        \caption{JointWM-WorldRec: incremental scene reconstruction via
                 scene fusion.}
        \label{fig:joint-worldrec}
    \end{subfigure}
    \hfill
    \begin{subfigure}[b]{0.48\linewidth}
        \centering
        \includegraphics[width=\linewidth]{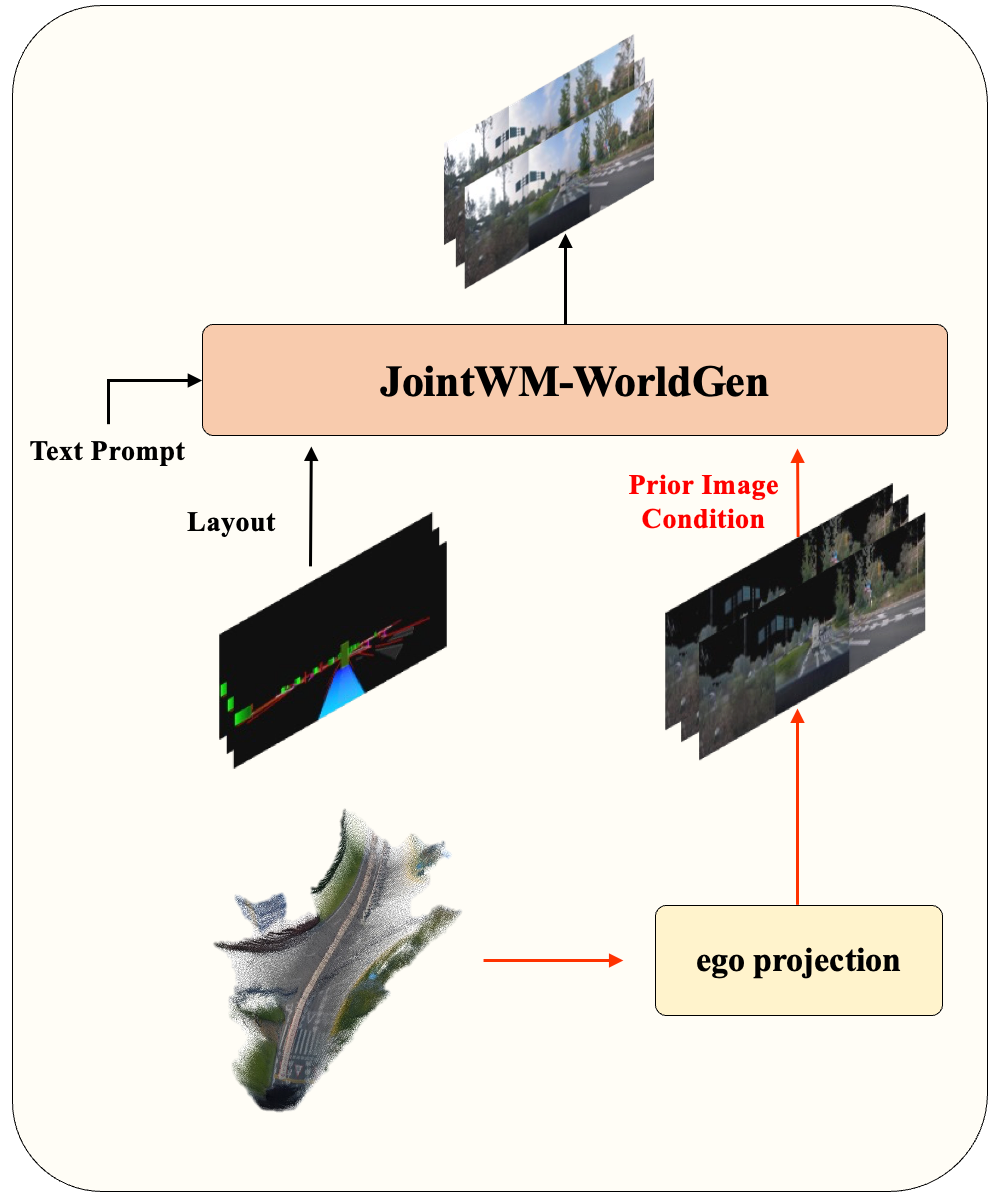}
        \caption{JointWM-WorldGen: RGB conditioning from ego-projected
                 rendered priors.}
        \label{fig:joint-worldgen}
    \end{subfigure}
    \caption{%
        Targeted adaptations of \textbf{WorldRec} and \textbf{WorldGen}
        for the \JWM (JointWM).
        (a) WorldRec fuses newly observed tokens with cached scene tokens to
        support incremental reconstruction.
        (b) WorldGen accepts an additional \emph{Prior Image Condition}
        rendered via ego projection from the reconstructed scene, providing a
        geometric scaffold for generation in unobserved regions.
    }
    \label{fig:joint-adaptations}
\end{figure}

To realize this tight coupling in practice, both \textbf{WorldRec} and
\textbf{WorldGen} are extended with targeted modifications that align their
interfaces and enable their cooperative operation within the \JWM.
Figure~\ref{fig:joint-adaptations} illustrates the two adapted modules.

\paragraph{WorldRec: incremental scene reconstruction.}
Standard feedforward reconstruction treats each input clip as an independent
unit, which limits the spatial extent of the recovered scene.
To better support the \JWM, we introduce a \emph{scene fusion} mechanism that
enables incremental reconstruction: given a scene already reconstructed from an
initial set of frames, \textbf{WorldRec} can ingest newly arriving images and
\emph{selectively update or extend} the existing Gaussian representation.
Specifically, newly observed tokens are fused with the cached scene tokens via a
cross-attention fusion layer, allowing the model to expand coverage into
previously unobserved regions, refine existing Gaussians with richer multi-view
evidence, and maintain a growing, globally consistent 4D scene representation as
the ego-vehicle traverses longer routes.
This incremental design is essential for closed-loop simulation, where the scene
must be continually extended as the vehicle enters new areas.

\paragraph{WorldGen: RGB conditioning from rendered priors.}
The original \textbf{WorldGen} conditions on ego trajectory and camera parameters.
Within the \JWM, we additionally introduce \emph{rendered-RGB conditioning}:
before generation, the scene tokens maintained by \textbf{WorldRec} are
rasterized into the target camera views, producing partial reference images that
may contain empty or occluded regions where the scene has not yet been observed.
These rendered priors are injected into the \textbf{WorldGen} DiT as an
additional conditioning modality, providing a coarse geometric scaffold that
guides synthesis in unobserved areas while preserving photometric consistency in
regions already covered by the reconstruction.
To support this conditioning scheme, we construct a dedicated training dataset
in which reference images are rendered from reconstructed scenes at held-out
target poses, then fine-tune \textbf{WorldGen} on this data.
The fine-tuning enables the model to robustly handle incomplete rendered
inputs---filling in missing content generatively while respecting the geometric
layout provided by the available rendered regions.

With these adaptations in place, the two modules operate as a tightly coupled
system in which each module's strengths compensate for the other's limitations.
The resulting \JWM achieves qualitative improvements that neither module can
deliver alone, which we characterize along three dimensions:
\begin{itemize}[leftmargin=*, itemsep=0.2em]
  \item \textbf{High stability:} Deterministic geometric constraints from
        \textbf{WorldRec} suppress error accumulation and content drift during
        long-horizon autoregressive generation.
  \item \textbf{High consistency:} The 4D scene representation serves as shared
        cross-frame memory, ensuring global consistency of object positions,
        lighting, and textures across different time steps and viewpoints,
        preventing hallucination artifacts.
  \item \textbf{High fidelity:} The generation model's rich conditioning signals,
        combined with strong supervision from real observations in the
        reconstruction module, bring synthesized content closer to real sensor
        observations, narrowing the simulation-to-real domain gap.
\end{itemize}

\section{Results}
\label{sec:results}

\subsection{WorldRec}
\label{sec:results-worldrec}

\begin{table}[t]
    \centering
    \caption{\textbf{Quantitative results on Waymo and nuScenes.}}
    \label{tab:worldrec_waymo}
    \resizebox{\columnwidth}{!}{%
        \begin{tabular}{lccccccc}
            \hline
            \multirow{2}{*}{Method} & \multicolumn{2}{c}{Waymo} & \multicolumn{2}{c}{NuScenes Zero-Shot} & \multicolumn{2}{c}{NuScenes Fine-Tuning}  \\
             & PSNR↑ & SSIM↑ & PSNR↑ & SSIM↑ & PSNR↑ & SSIM↑ \\ \hline

            MVSSplat \cite{chen2024mvsplat}&   20.56 & 0.697 &  17.84 & 0.563 &  - & - \\ 
             NoPoSplat \cite{ye2024noposplat} & 24.31 & 0.751 &  19.75 & 0.545 &  - & -\\ 
             DepthSplat  \cite{Xu_2025_CVPR} & 23.26 & 0.696 &  19.52 & 0.601&  - & - \\ 
            STORM \cite{yang2025storm} & 26.38 & 0.794 & 17.77 & 0.669 & 24.54 & 0.784  \\
            DGGT \cite{chen2025dggt} & 27.41  & 0.846  & 25.31 & 0.794 & 26.63 & 0.813  \\
            Ours & \textbf{28.48 } & \textbf{0.861}  & \textbf{26.54} & \textbf{0.821} & \textbf{27.50} & \textbf{0.826} \\ \hline
        \end{tabular}%
    }
\end{table}

\begin{figure}[htbp]
    \centering
    \includegraphics[width=\linewidth]{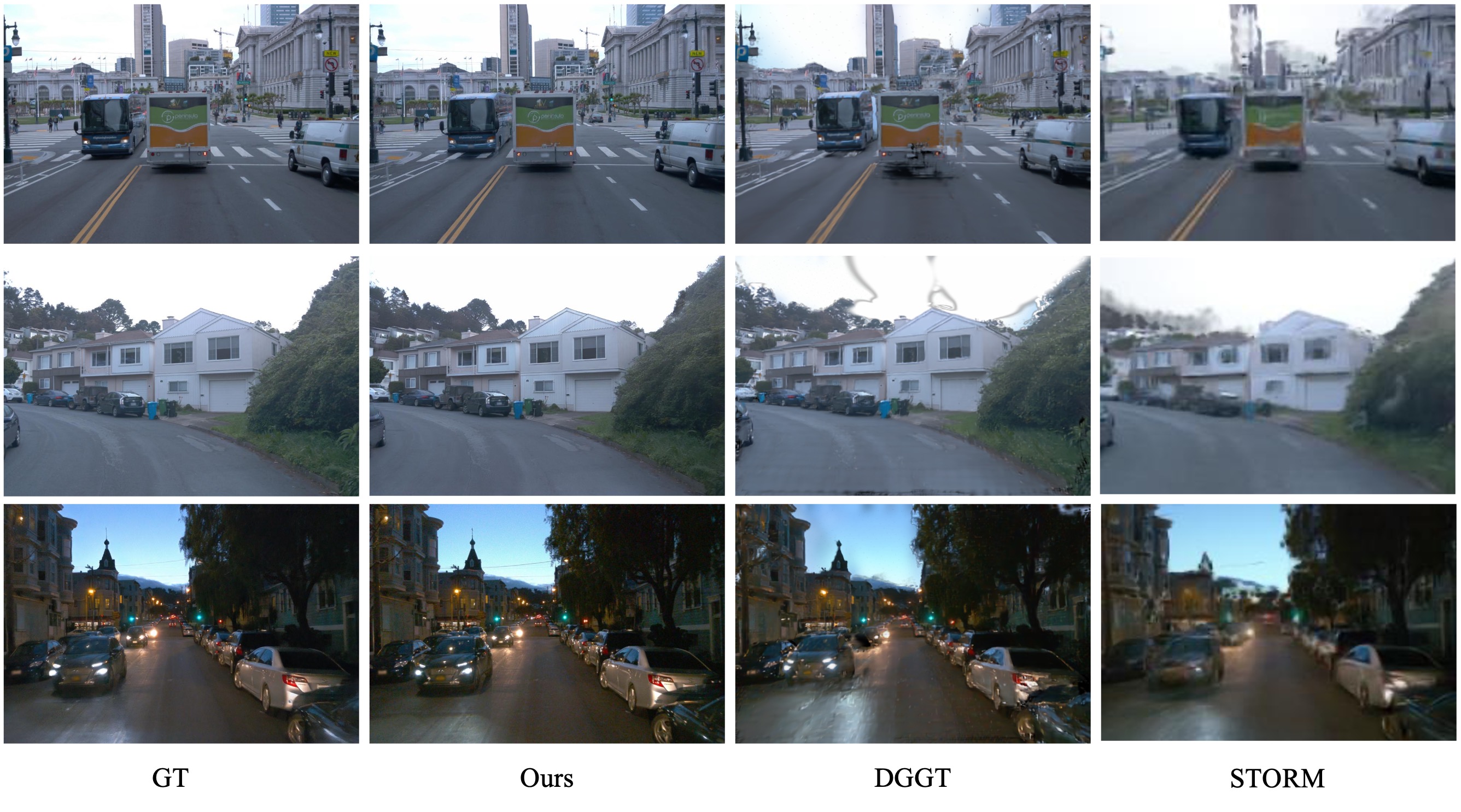}
    \caption{%
     \centering
      Scene reconstruction results on original trajectories.
    }
    \label{fig:worldrec-orig}
\end{figure}

\begin{figure}[htbp]
    \centering
    \includegraphics[width=\linewidth]{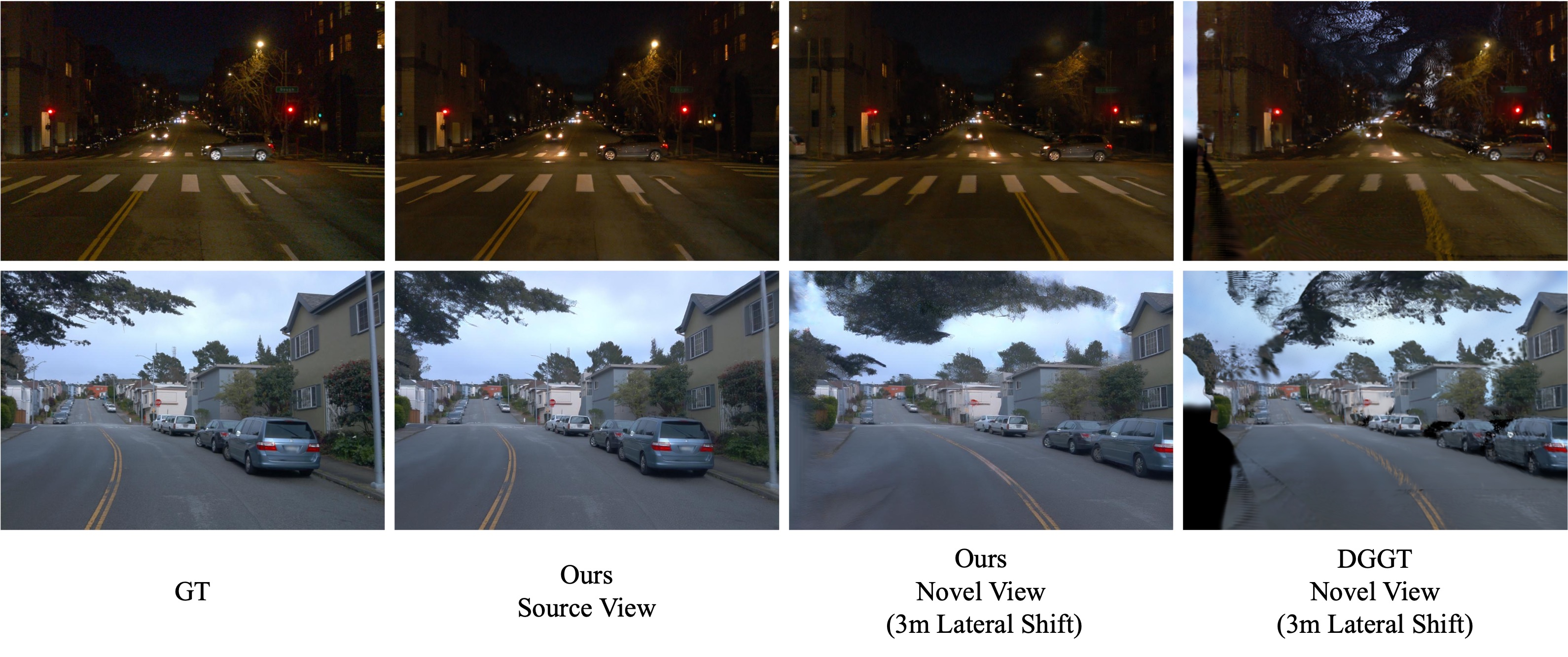}
    \caption{%
     \centering
      \textbf{WorldRec} novel view synthesis on Waymo.
    }
    \label{fig:nvs-waymo}
\end{figure}

\begin{figure}[htbp]
    \centering
    \includegraphics[width=\linewidth]{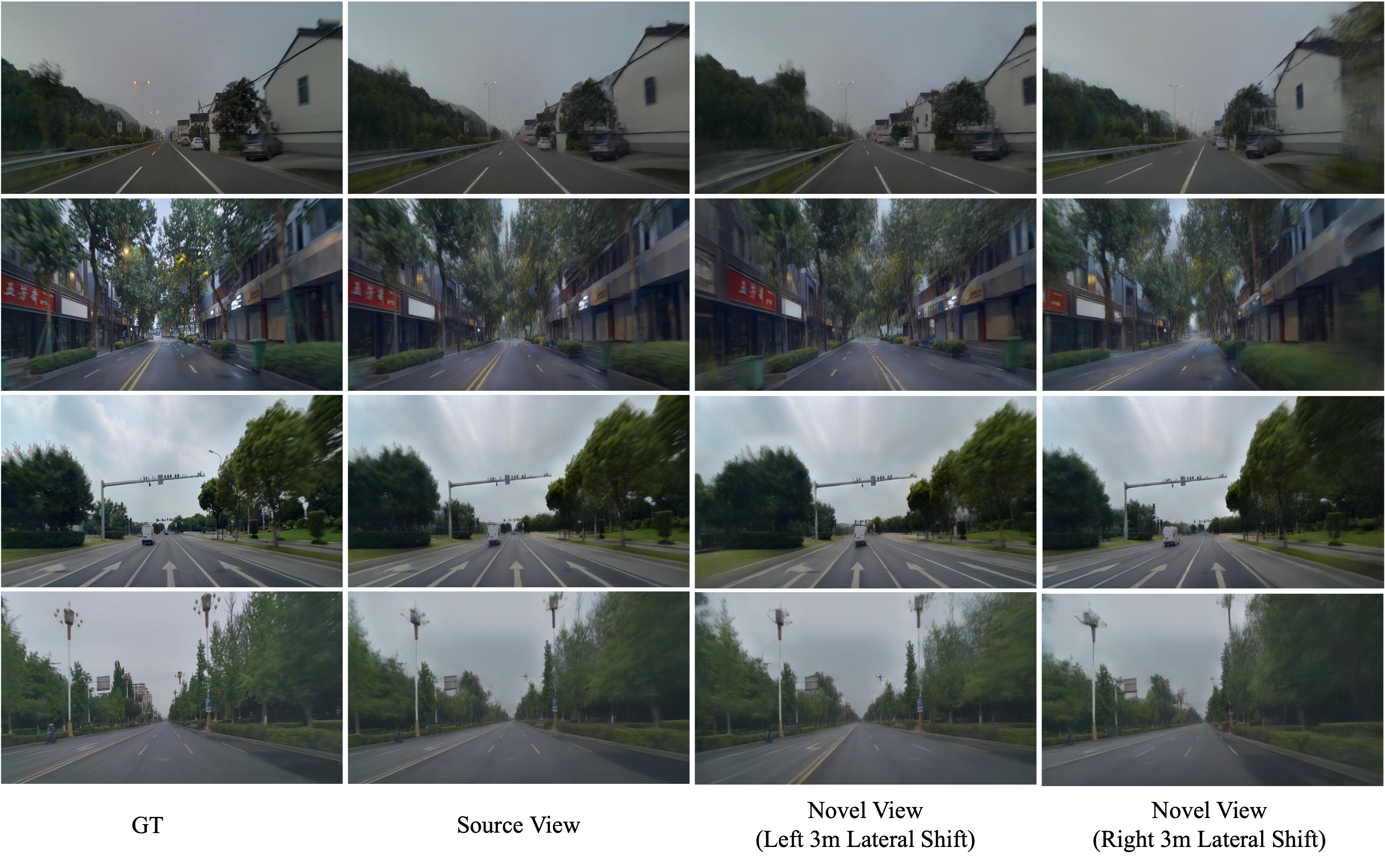}
    \caption{%
     \centering
      \textbf{WorldRec} novel view synthesis on private data.
    }
    \label{fig:nvs-mi}
\end{figure}

\begin{figure}[htbp]
    \centering
    \includegraphics[width=\linewidth]{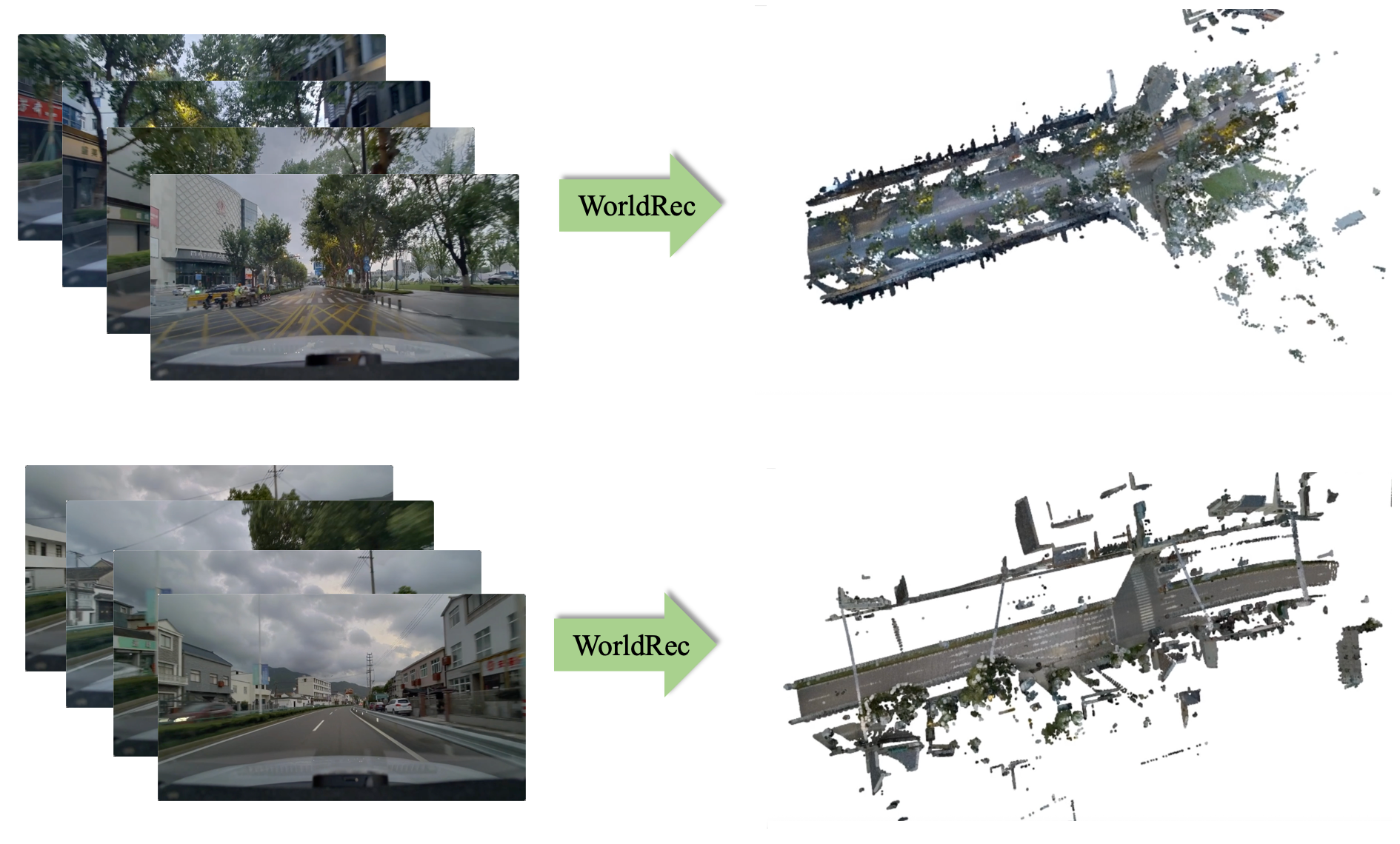}
    \caption{%
     \centering
      \textbf{WorldRec} bird's-eye view reconstruction on private data.
    }
    \label{fig:worldrec-bev}
\end{figure}

Table \ref{tab:worldrec_waymo} presents a quantitative comparison of our method against current state-of-the-art approaches on the Waymo and nuScenes benchmarks, demonstrating that our method achieves superior performance across both datasets. Figure~\ref{fig:worldrec-orig} shows reconstruction results on original driving
trajectories on the Waymo dataset~\cite{sun2020scalability}.
Figures~\ref{fig:nvs-waymo} and~\ref{fig:nvs-mi} present novel view synthesis quality on Waymo
and private data respectively. These results demonstrate the clear superiority of our method over state-of-the-art baselines in both scene reconstruction and novel view synthesis.
Figure~\ref{fig:worldrec-bev} further illustrates bird's-eye view reconstruction on private data,
demonstrating that \textbf{WorldRec} produces geometrically coherent and spatially complete
scene representations beyond the ego-vehicle's forward-facing cameras,
validating its generalization to diverse sensor configurations and scene layouts.

Beyond reconstruction quality, our method also enables highly efficient scene reconstruction. For a 10-second video clip, our method completes the reconstruction in roughly 10 seconds, whereas per-scene optimization takes around 4 hours.

\subsection{WorldGen}
\label{sec:results-worldgen}

\begin{figure}[htbp]
    \centering
    \includegraphics[width=\linewidth]{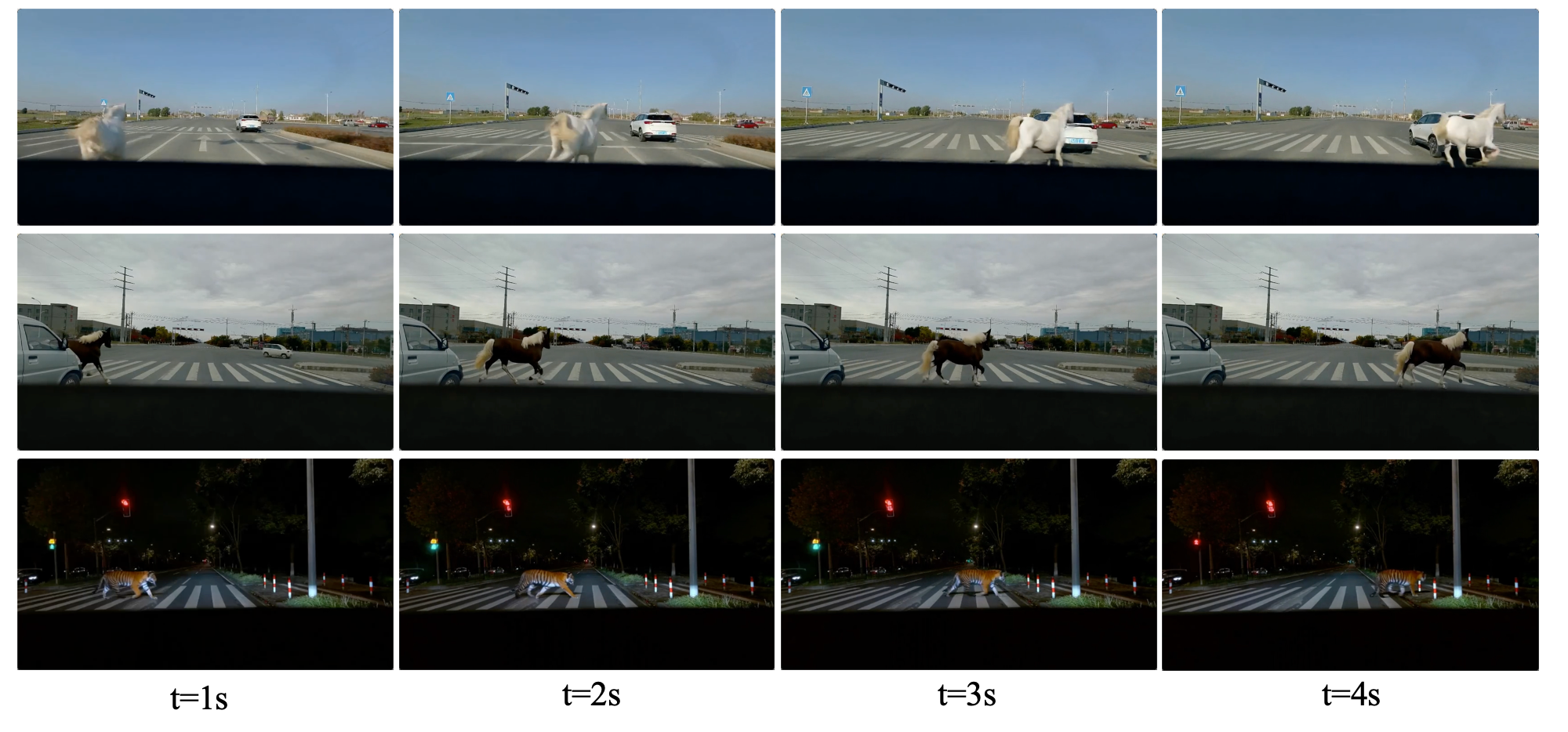}
    \caption{%
     \centering
      \textbf{WorldGen} long-tail scene generation: animals on road.
    }
    \label{fig:gen-longtail}
\end{figure}

\begin{figure}[htbp]
    \centering
    \includegraphics[width=\linewidth]{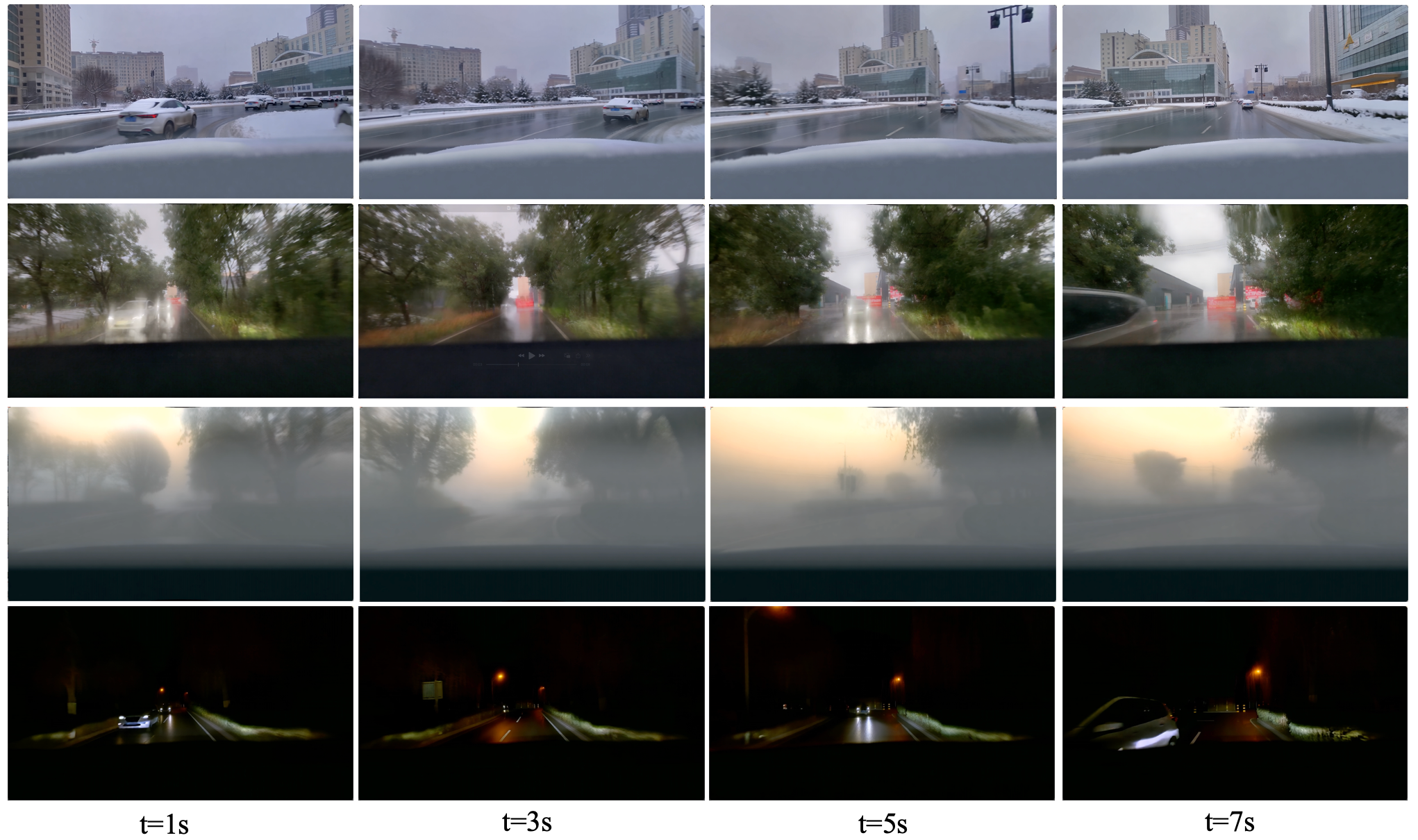}
    \caption{%
     \centering
      \textbf{WorldGen} extreme weather scene generation.
    }
    \label{fig:gen-weather}
\end{figure}


\begin{figure}[htbp]
    \centering
    \includegraphics[width=\linewidth]{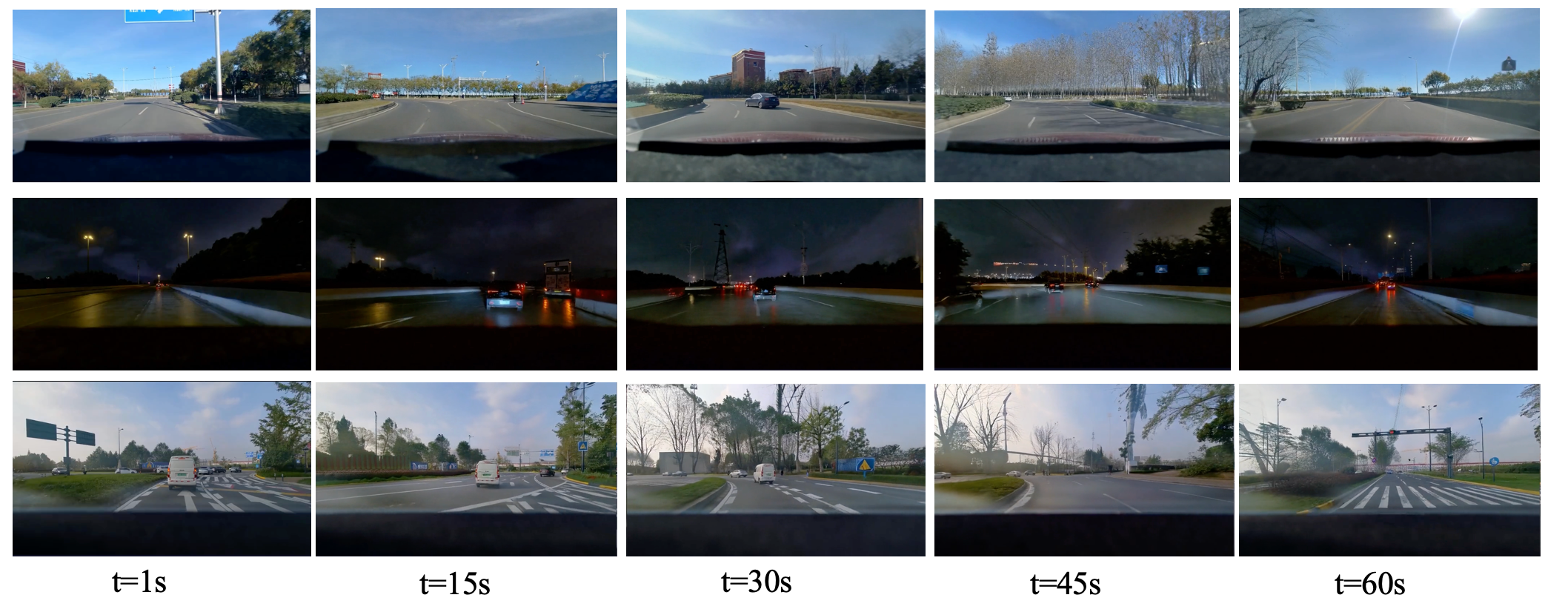}
    \caption{%
     \centering
      \textbf{WorldGen} controllable long-horizon generation (10fps/30\,fps, $\leq\!1$\,min).
    }
    \label{fig:gen-control}
\end{figure}

Figures~\ref{fig:gen-longtail}--\ref{fig:gen-control} present qualitative generation results across
four challenging scenario categories.
\textit{Long-tail animal scenes} (\figref{fig:gen-longtail}) demonstrate the
model's ability to synthesize rare events such as tigers and horses intruding
on the road.
\textit{Extreme weather} (\figref{fig:gen-weather}) shows high-fidelity
generation of adverse conditions including rain, snow, and fog.
\textit{Controllable long-horizon generation} (\figref{fig:gen-control})
confirms temporal stability at 30\,fps for sequences up to one minute,
enabled by the 4-step efficient sampling pipeline.
On H20 GPUs, \textbf{WorldGen} generates at 0.19\,s/frame (single view) and
0.46\,s/frame (three views).
Tab.~\ref{tab:comparison} compares our method against representative driving world models on the nuScenes dataset, covering both bidirectional (Bi) and autoregressive (AR) approaches. Our method, as an autoregressive model, achieves an FID of 7.04 and FVD of 64.97, outperforming all listed models in FVD while maintaining competitive FID. Notably, our approach generates significantly longer videos of 81 frames, far exceeding the 8--16 frames produced by most baselines. Compared to the only other AR method, Epona, our model achieves a lower FVD (64.97 vs.\ 82.8) with a substantially faster inference time of 0.19\,s versus 1.06\,s per frame, demonstrating both superior generation quality and efficiency.


\begin{table*}[t]
\centering
\caption{Comparison of driving world models on nuScenes dataset.}
\label{tab:comparison}
\resizebox{\textwidth}{!}{
\begin{tabular}{lcccccc}
\toprule
\textbf{Model} & \textbf{Bi or AR} & \textbf{Venue} & \textbf{FID $\downarrow$} & \textbf{FVD $\downarrow$} & \textbf{Frames} & \textbf{Infer. Time} \\
\midrule
MagicDrive~\cite{magicdrive} & Bi & ICLR'24 & 16.20 & - & 1 & - \\
MagicDrive-V2~\cite{magicdriveV2} & Bi & ICCV'25 & 20.91 & 94.84 & 16 & - \\
Vista~\cite{vista} & Bi & NeurIPS'24 & 6.9 & 89.4 & 16 & - \\
DiVE~\cite{jiang2025diveefficientmultiviewdriving} & Bi & arXiv'25 & 7.14 & 68.4 & 8 & - \\
Delphi~\cite{ma2024unleashinggeneralizationendtoendautonomous} & Bi & arXiv'24 & 15.08 & 113.5 & 8 & - \\
UniScene~\cite{li2024uniscene} & Bi & CVPR'25 & 6.12 & 70.52 & 8 & - \\
Genesis~\cite{genesis} & Bi & NeurIPS'25 & 6.45 & 67.87 & 16 & - \\
Epona~\cite{zhang2025eponaautoregressivediffusionworld} & AR & ICCV'25 & 7.5 & 82.8 & 16 & 1.06 \\
Ours~ & AR & - &  7.04 & 64.97 & 81 & 0.19 \\
\bottomrule
\end{tabular}
}
\end{table*}

\subsection{Joint World Model}
\label{sec:results-joint}

We evaluate the \JWM along three complementary dimensions: long-horizon temporal
consistency, multi-view spatial consistency, and multi-run stability.

\begin{figure}[htbp]
    \centering
    \includegraphics[width=\linewidth]{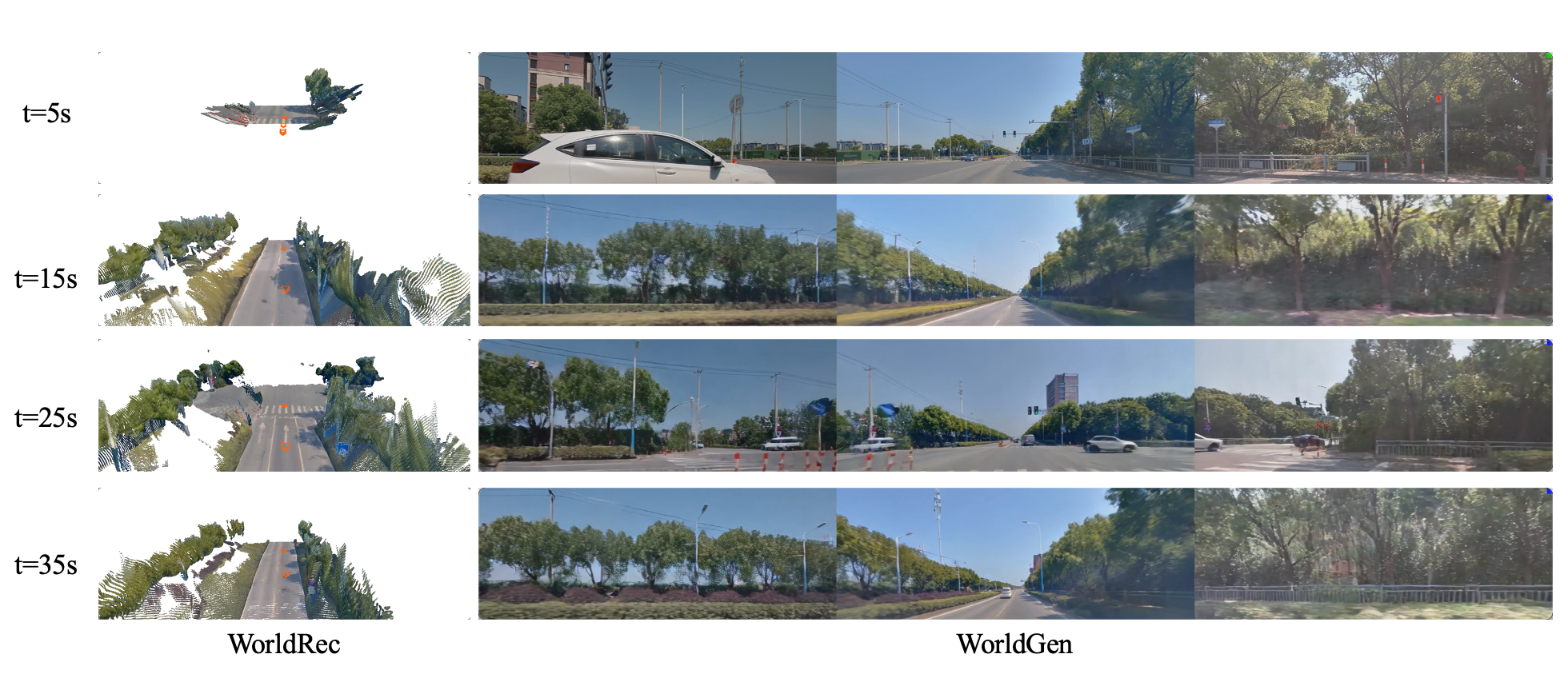}
    \caption{%
      \JWM long-horizon temporal consistency (example 1). Scene geometry and
      dynamic content remain stable over extended generation horizons without
      drift or hallucination.
    }
    \label{fig:jwm-1}
\end{figure}

\begin{figure}[htbp]
    \centering
    \includegraphics[width=\linewidth]{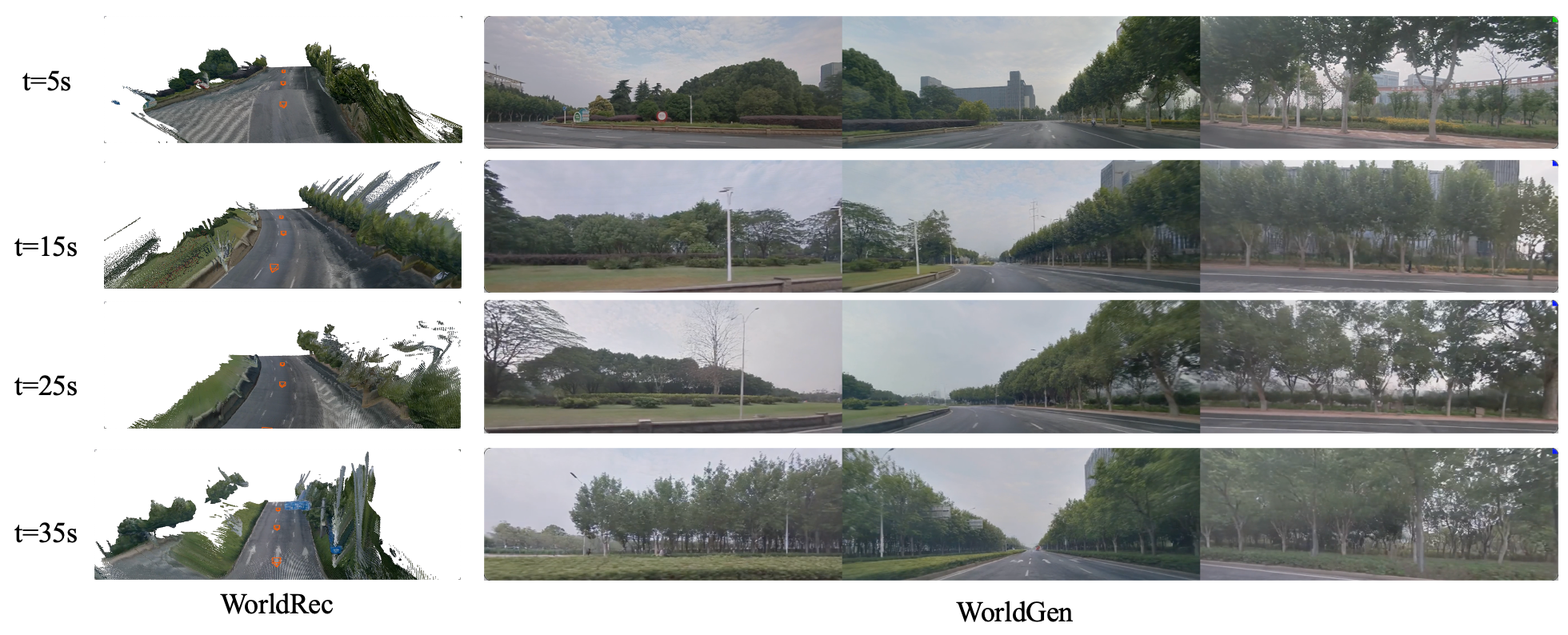}
    \caption{%
      \JWM long-horizon temporal consistency (example 2). The geometric prior
      from \textbf{WorldRec} anchors the generative trajectory of
      \textbf{WorldGen}, preserving structural coherence across the full
      sequence.
    }
    \label{fig:jwm-2}
\end{figure}

\begin{figure}[htbp]
    \centering
    \includegraphics[width=\linewidth]{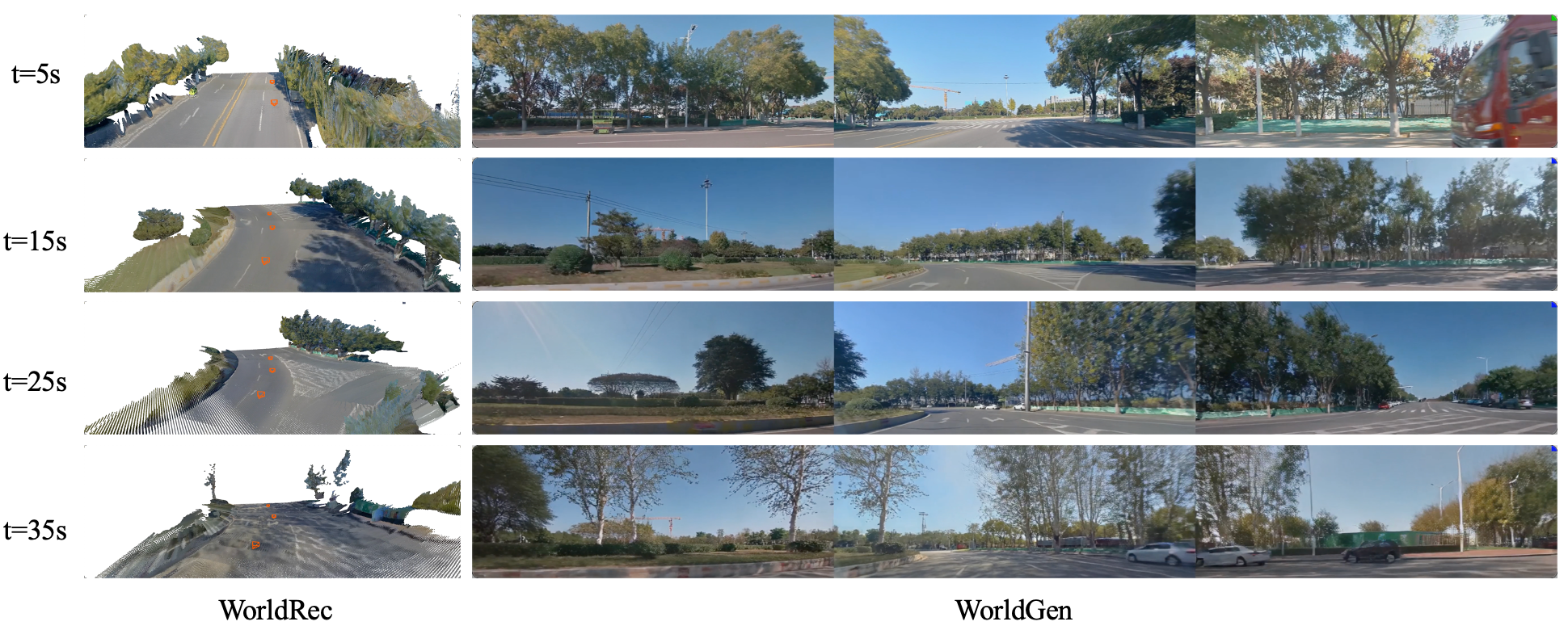}
    \caption{%
      \JWM long-horizon temporal consistency (example 3). Consistent road
      structure, lighting, and dynamic actor positions are maintained throughout
      the generated sequence.
    }
    \label{fig:jwm-3}
\end{figure}

\paragraph{Long-horizon temporal consistency.}
A key challenge in autoregressive generation is the accumulation of errors over
time, which leads to content drift and visual degradation in long sequences.
As shown in Figures~\ref{fig:jwm-1}, \ref{fig:jwm-2}, and~\ref{fig:jwm-3},
the \JWM maintains coherent scene structure throughout extended generation
horizons.
The deterministic geometric prior supplied by \textbf{WorldRec} anchors the
generative process of \textbf{WorldGen}, preventing the drift and hallucination
artefacts that appear in standalone generation baselines.
Scene elements such as lane markings, road boundaries, and dynamic objects
remain geometrically stable across frames, even when the generation window
extends well beyond the observed input.

\begin{figure}[htbp]
    \centering
    \includegraphics[width=\linewidth]{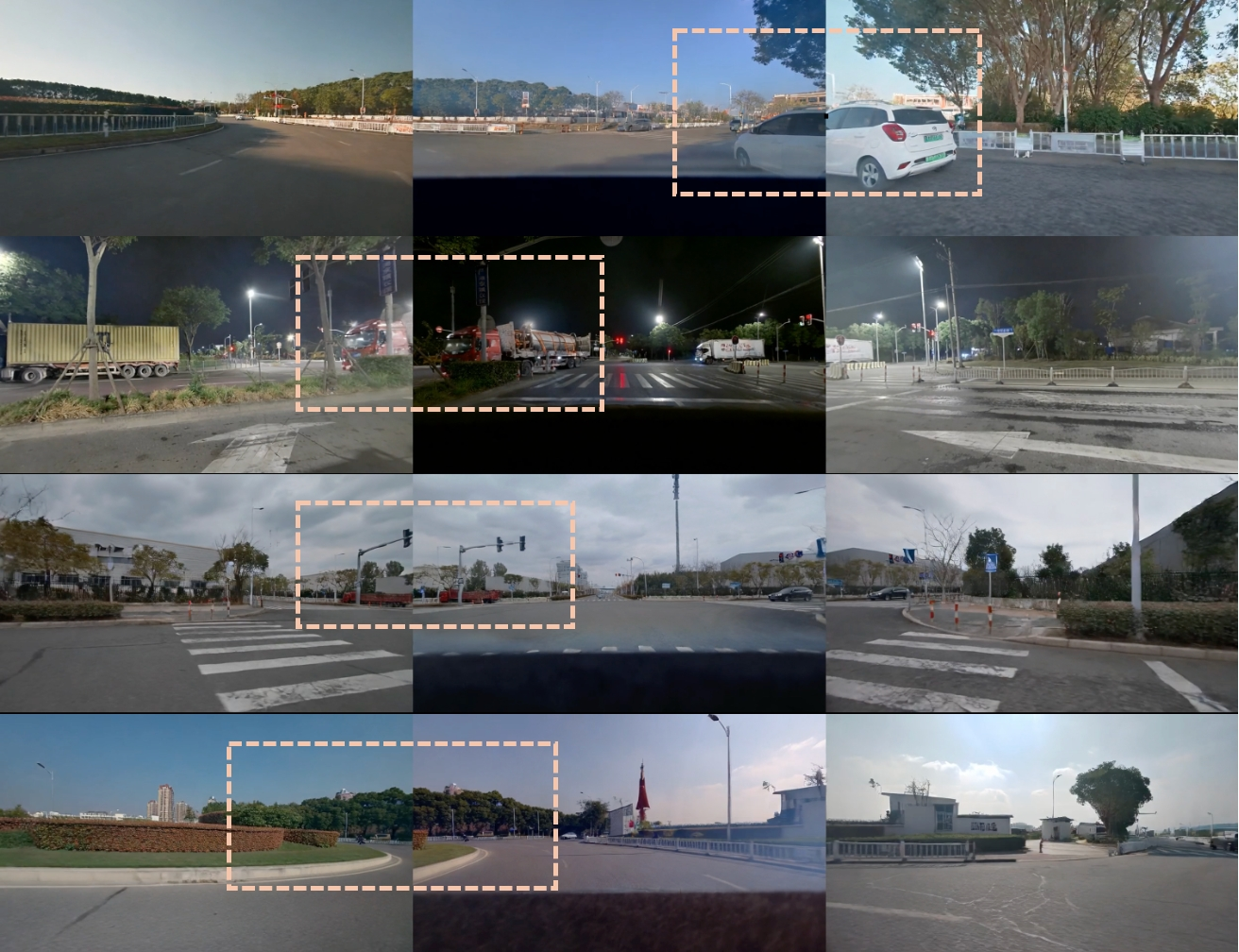}
    \caption{%
      \JWM multi-view spatial consistency. 
    }
    \label{fig:jwm-4}
\end{figure}

\paragraph{Multi-view spatial consistency.}
Maintaining consistent appearance across simultaneously rendered camera views
is essential for downstream perception and simulation tasks.
\figref{fig:jwm-4} illustrates multi-view outputs from the \JWM, showing that
the 4D scene representation produced by \textbf{WorldRec} acts as shared spatial
memory across all camera viewpoints.
Consequently, object positions, lighting conditions, and surface textures are
globally coherent across the full camera rig, eliminating the cross-view
inconsistencies and hallucination artefacts that arise when each view is
generated independently.

\begin{figure}[htbp]
    \centering
    \includegraphics[width=\linewidth]{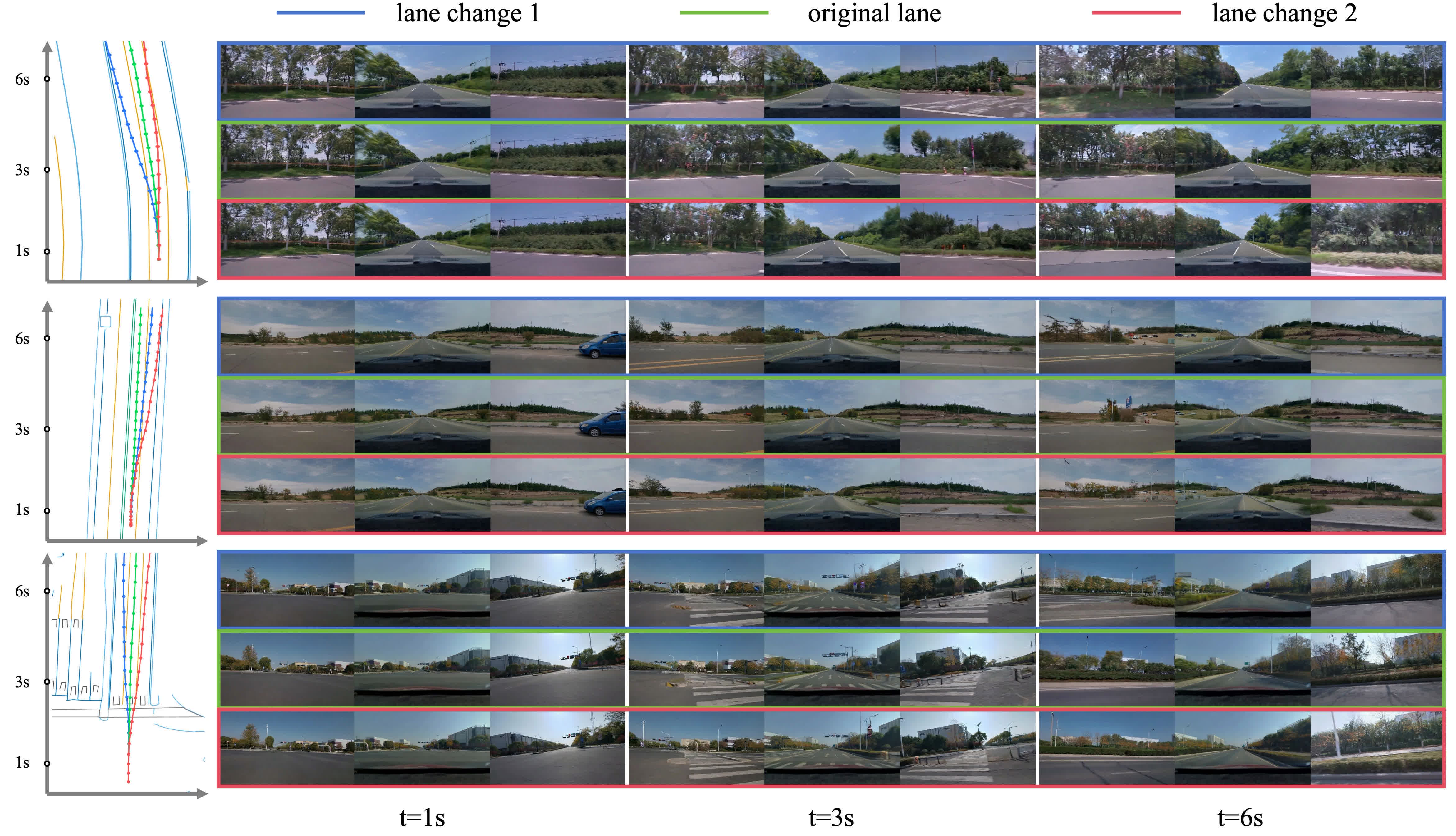}
    \caption{%
      \JWM multi-run stability. Repeated inference under identical conditions
      yields structurally consistent outputs, demonstrating that
      \textbf{WorldRec}'s geometric constraints reduce generative variance.
    }
    \label{fig:jwm-5}
\end{figure}

\paragraph{Multi-run stability.}
Generative models can exhibit high variance across different inference runs,
complicating reproducible simulation and closed-loop evaluation.
\figref{fig:jwm-5} demonstrates the multi-run stability of the \JWM:
repeated generation under identical conditions produces structurally consistent
outputs, owing to the deterministic geometric constraints imposed by
\textbf{WorldRec}.
This stability is critical for reliable data synthesis and for fair comparison
in closed-loop evaluation pipelines.

\section{Conclusion}
\label{sec:conclusion}

This report has systematically presented the technical designs and experimental results of \textbf{WorldRec}, \textbf{WorldGen}, and the \textbf{\JWM}---which arises from the deep integration of the former two---built around the two core capabilities of world models for autonomous driving.

\textbf{WorldRec} breaks through two long-standing bottlenecks---the multi-hour
per-scene optimization cost and the Gaussian primitive explosion of per-pixel
feed-forward methods---by adopting a sparse-query-driven feed-forward
reconstruction paradigm.
It compresses reconstruction time to seconds while maintaining high-fidelity
rendering, laying the foundation for large-scale engineering deployment.

\textbf{WorldGen} adopts a two-stage strategy of bidirectional pretraining followed
by causal fine-tuning, balancing generation quality and inference efficiency:
bidirectional pretraining fully learns the global scene distribution, and
three-stage causal fine-tuning (Teacher Forcing $\to$ ODE distillation $\to$
DMD) progressively resolves the causal constraint, inference latency, and
exposure bias challenges, ultimately achieving high-quality long-horizon video
generation with only 4 denoising steps.

The \textbf{\JWM} organically integrates both modules: the deterministic geometric constraints from WorldRec suppress generative drift, while the rich imagination of WorldGen compensates for the limitations of reconstruction-only methods in unseen regions, achieving synergistic improvements along three axes---stability, consistency, and fidelity.

In summary, the technical system presented in this report provides a complete
solution for constructing high-quality autonomous driving world models suitable
for closed-loop simulation, data synthesis, and end-to-end training, with strong
theoretical value and practical engineering significance.

\phantomsection
\addcontentsline{toc}{section}{Contributors}  
\section*{Contributors}
\label{sec:contributors}

\noindent
\textbf{Reconstruction Model:} Hongcheng Luo $\cdot$ Cheng Chi $\cdot$ Mingfei Tu $\cdot$ Lei Gong \\[4pt]
\textbf{Generation Model:} Lijun Zhou $\cdot$ Zhenxin Zhu $\cdot$ Zhanqian Wu $\cdot$ Kaixin Xiong\\[4pt]
\textbf{Joint World Model:} Lijun Zhou $\cdot$ Mingfei Tu $\cdot$ Hongcheng Luo $ \cdot$ Kaixin Xiong \\[4pt]
\textbf{Project Advisor:} Guang Chen $\cdot$ Hangjun Ye \\[4pt]
\textbf{Project Lead:} Haiyang Sun $\cdot$ Bing Wang

\section*{Acknowledgements}
\label{sec:acknowledgements}

We sincerely thank the following colleagues for their valuable contributions and
support:
Zehan Zhang $\cdot$
Fangzhen Li $\cdot$
Hao Li $\cdot$
Yingying Shen $\cdot$
Jiale He $\cdot$
Haohui Zhu $\cdot$
Shan Zhao $\cdot$
Kai Wang $\cdot$
Zhiwei Zhan $\cdot$
Yuechuan Pu $\cdot$
Kaiyuan Tan $\cdot$
Ruiling Yang $\cdot$
Xianqi Wang $\cdot$
Tianyi Yan $\cdot$
Jiawei Zhou $\cdot$
Lei Zhang $\cdot$
Jingyang Zhao $\cdot$
Xi Zhou $\cdot$
Chitian Sun $\cdot$
Chenming Wu $\cdot$
Jiong Deng $\cdot$
Hongwei Xie $\cdot$
Ming Lu $\cdot$
Kun Ma $\cdot$
Long Chen.

{
    \small
    \bibliographystyle{ieeenat_fullname}
    \bibliography{paper}
}

\end{document}